\documentclass[sigconf]{acmart}

\usepackage[utf8x]{inputenc}
\usepackage[english]{babel}
\usepackage{url}
\usepackage{lipsum}
\usepackage{listings}
\usepackage{graphicx}
\usepackage{color}
\usepackage{amsmath}
\usepackage[]{algorithm2e}
\usepackage{mathtools}
\usepackage{algorithmic}

\usepackage{hyperref}

\def\ls{\left(}
\def\rs{\right)}

\def\mera{\mathsf{P}}

\def\me{\mathsf{E}}

\AtBeginDocument{%
  \providecommand\BibTeX{{%
    \normalfont B\kern-0.5em{\scshape i\kern-0.25em b}\kern-0.8em\TeX}}}

\setcopyright{acmcopyright}
\copyrightyear{2020}
\acmYear{2020}
\acmDOI{}

\acmConference[Preprint]{}{}{}
\acmBooktitle{}
\acmPrice{}
\acmISBN{}

\begin{document}

\title{COHORTNEY: Non-Parametric Clustering of Event Sequences}

\author{Vladislav Zhuzhel}
\authornote{Equal contribution}
\author{Rodrigo Rivera-Castro}
\authornotemark[1]
\authornote{E-mail: rodrigo.riveracastro@skoltech.ru}
\affiliation{%
  \institution{Skoltech}
  \city{Moscow}
  \country{Russia}
}

\author{Nina Kaploukhaya}
\affiliation{
  \institution{Skoltech, IITP RAS, Moscow Institute of Physics and Technology}
  \city{Moscow}
  \country{Russia}
}

\author{Liliya Mironova}
\affiliation{%
  \institution{SKoltech}
  \city{Moscow}
  \country{Russia}
  }

\author{Alexey Zaytsev}
\authornotemark[1]
\authornote{E-mail: a.zaytsev@skoltech.ru}
\author{Evgeny Burnaev}
\affiliation{%
  \institution{Skoltech}
  \city{Moscow}
  \country{Russia}}

\renewcommand{\shortauthors}{Rivera-Castro, et al.}

\begin{abstract}
Cohort analysis is a pervasive activity in web analytics. 
One divides users into groups according to specific criteria and tracks their behavior over time. 
Despite its extensive use, academic circles do not discuss cohort analysis to evaluate user behavior online. 
This work introduces an unsupervised non-parametric approach to group Internet users based on their activities. 
In comparison, canonical methods in marketing and engineering-based techniques underperform. 
COHORTNEY is the first machine learning-based cohort analysis algorithm with a robust theoretical explanation.
\end{abstract}




\maketitle
\section{Introduction}

Event sequences are commonplace in practice. 
We can look at some illustrative examples of such events.

\subsection{Examples of event sequences}

\paragraph{Triggers in databases}
A trigger is a special type of procedure that is executed when a specific event occurs
We can implement triggers through constant queries to the corresponding table. 
Each entry in the activity table displays an event that occurred with the database.
It can be, for example, adding, deleting, or changing records. 
Each new event, i.e. each new entry in the activity table, triggers a trigger.
We use triggers to ensure data integrity and implement complex business logic. 
The triggers modify specific transactions in the data.
Accordingly, if we detect an error or data integrity violation, we can roll back the corresponding transaction.

\paragraph{Model of the insurance fund}
An example of an event sequence is the model for the social security fund. 
The main characteristic of the fund is its capital $S(t)$ at the time of $t$. 
The following changes may occur with this capital:
First, the fund receives funds from businesses and organizations. 
We can assume that they arrive continuously in time at the speed of $c_0$.
Second, an insurer makes payments to the fund. 
We assume that the sequence of insurance payments is a Poisson process with constant intensity, $\lambda$.
The insurance payments themselves, $\xi$, are independent equally distributed random variables with exponential distribution

\begin{equation*}
    p_{\xi}(x) = {\frac 1 a}e^{-\frac x a}, x\ge0
\end{equation*}

Third, the institution allocates part of its funds to social programs. 
In this model, we assume that the process of allocating money for social needs forms a Poisson process of variable intensity, $\mu(S)$.
Moreover, the payments, $\eta$, are independent and equally distributed random variables with exponential distribution.

\paragraph{Broker lending to its clients}
A broker lending to its clients is a process.
We can represent this process as a queuing system.
Moreover, we can understand it as a set of servicing devices and serviced requirements, for example, the requests and orders, from a particular incoming sequence of conditions.
We assume that there is a sequence of client requests coming to the broker with an intensity, $A$.
Further, there are no restrictions on the number of these requests. 
We can carry out the maintenance at various intensities.
The set of values of which is finite, for example, $B(1), B(2), \ldots B(K)$. 
We number these service intensities in ascending order, i.e. $B(1) < B(2) < … < B(K)$, with $A < B(i),\forall i=\overline{1,K} $. 
In other words, it is always possible to increase the service intensity, so that it overlaps the intensity of incoming requests. 
Service durations at all intensities are mutually independent and exponentially distributed random variables.
They do not depend on the incoming sequence of requests.
For each unit of time spent with the broker of $n$ orders, we charge a penalty $C(n)\ge 0$, $n = 0, 1, 2,\ldots$. 
The function $C (n)$ does not decrease by $n$ and $C(n)\rightarrow\infty$ for $n\rightarrow\infty$.
For each unit of time during which we use the $i$ intensity, we charge a penalty, $h(i)$. 
We assume that the penalty increases with increasing intensity, i.e. $h(i)\le h(i+1), \forall i=\overline{1,K-1}$.
Moreover, we manage the service mechanism.
Hence, we assign the service intensity to use when there are $n>0 $ requests.
The criterion is the average cost of the broker's operation per unit of time in stationary mode.
We define it as

\begin{equation*}
    U = C(0)P_0 + \sum_n{P_n[C(n) + h(k(n))]}.
\end{equation*}

Here, $P_n$ is the stationary probability of finding $n$ orders at the broker.
$k(n)$ is the number of the service intensity that we use when the broker has $n$ orders.
As we can see from the examples, we can find event sequences in any field of human activity. 

\subsection{Mathematical formalization of event sequences}
Let there be a sequence of events $a_1, a_2, \ldots$. 
We assign to each event from the sequence a numeric characteristic, $A$.
It depends on all events up to the current time, $A(a_1, a_2, \ldots a_N)\to~R$.
Here $N$ is the current time.
The modelling of event sequences varies depending on how we describe the numerical characteristics of $A$.

\paragraph{The time series model}
At each time, $i$, a value, $A[i]$, describes the event, $a_i$. 
It is a suitable model when, for example, there is traffic passing through an IP node every 5 minutes.
Another example is for the level of trading on the exchange every minute.
This model is also appropriate for describing triggers in a database when we record events occurring at specific time points.

\paragraph{Model of the cash register}
In this model, when the event, $a_i$, occurs, the value is $A[j]$. 
We let $a_i = (j, I_i), I_i \ge 0, A_i[j] = A_{i-1}[j] + I_i$ increase.
Here, $j\in J$ is a type of numeric characteristic, $A_i$ is the value after receiving $i$-th events in the sequence. 
It is probably the most popular data sequence model. 
It is well suited for monitoring IP addresses when accessing a web server and sending data packets.
The same IP addresses may visit the server or send packets multiple times over a period.

\paragraph{The turnstile model}
Here $a_i$ shows the change in $A[j]$. 
For example, we have $a_i = (j, U_i)$, $A_i[j] = A_{i-1}[j] + U_i$.
In this case, $j\in J$ is a type of numeric characteristic.
$A_i$ is the value of the signal after we receive the event, $i$, in the sequence.
$U_i$ can be either positive or negative. 
It is the most common model. 
An example is a subway where people both arrive at the station and leave it through turnstiles, and there is always high traffic in the metro. 
It is the most appropriate model to describe a fully dynamic situation where both inserts and deletions are possible.
However, it is always challenging to get boundaries of interest in this model.
In some cases, it is $A_i[j] \ge 0$ for all $i$. 
In this case, we talk about a strict turnstile model. 
In practice, this corresponds to the fact that people are the only exit through the turnstiles through which they enter.
It looks unrealistic, but it has many applications. 
For example, in a database, we can delete only our records. 
On the other hand, there are examples where data sequences can be non-strict, i.e. $A_i[j]<0$ for some $i$.
This model is also suitable to describe the capital of an insurance fund, which can either increase due to the receipt of new funds or decrease through insurance payments.
We can see that the turnstile model is the most common.
In contrast, the cash register model is a particular case of the turnstile model. 
From a theoretical point of view, of course, it is convenient to design different algorithms using the most general model, the turnstile model.
From a practical point of view, a weaker and less general model is easier to use for different applications.

\section{Forecasting Event Sequences}
We use the cash register model for further description.
Let there be some event sequence, $c = (a_1, a_2 \ldots a_k)$, and some specified number, $\alpha$. 
The task is to specify a time interval, $\tau$, such that $P(A_{k+1}[j] \in [A_k[j], A_k[j] + \tau]\mid {\mathcal{F}}_t) > \alpha$. 
That is, we must learn how to predict the next event as accurately as possible.
For this, we use the history of events up to the current time.
The mathematical statement of the problem is as follows. 
We let $\theta$ be a moment in time from the interval $(t,\infty)$ when another event  appears in the sequence under consideration, i.e. $N_ {(t,\theta]}=1$.
We denote the predicted time of occurrence of a new event as $\tau\geq t$.
We look for a moment in time, $\tau^*\in (t,\infty]$
It must achieve a minimum in

\begin{equation}\label{eq1}
    S(\tau,c)=P\left(\tau\leq\theta\mid {\mathcal{F}}_t\right) + c\mathsf{E} \left(G\left([\tau-\theta]^+\right)\mid {\mathcal{F}}_t\right).
\end{equation}

We define $c>0$ as some constant that characterizes "importance" to minimize the delay in detecting a new event, $(x)^+=\max(x,0)$.
Further, $G = G(x)$ is some non-decreasing penalty function, with $G (x)\geq0$ for $x\geq0$, and $G(0)=0$.
The event history is ${\mathcal{F}}_t$, i.e. the information about events that occurred up to the time $t$.
As a function of $G(x)$, we can use two types of functions.
First, we can consider a linear function.
One example is $G(x) = x$.
Second, we can take a nonlinear function that satisfies four conditions.
The function must be $G(0) = 0$.
For $x>0$, we must have $G(x)\ge 0$.
$G(x)$ must increase when $x>0$.
On the limit, it satisfy $\lim_{x\rightarrow\infty}G(x) = const < \infty$.
Thus, by solving the optimization problem (\autoref{eq1}), we can get the optimal forecast for $\tau^*$.
To do this, we calculate the conditional probability $P\left(\tau\leq\theta\mid {\mathcal{F}}_t\right)$ and the conditional expectation $\mathsf{E} \left(G\mid {\mathcal{F}}_t\right)$.
There are two approaches to define the former, $P\left(\tau\leq\theta\mid {\mathcal{F}}_t\right)$, and the latter, $\mathsf{E} \left(G\mid {\mathcal{F}}_t\right)$.

\textbf{Method 1 (parametric).}
Based on information from a domain area, we define a parametric distribution of events $a_1 \ldots a_k$. 
For example, we assume that the time intervals $(a_k - a_{k-1})$ between two neighbouring events have a Poisson distribution with variable intensity. 
Based on the available data, we evaluate the parameters of this distribution.
Then, we calculate $P\left(\tau\leq\theta\mid {\mathcal{F}}_t\right)$, $\mathsf{E} \left(G\mid {\mathcal{F}}_t\right)$.

\textbf{Method 2 (nonparametric).}
We estimate the conditional probability and mathematical expectation based on an existing subset of close, in a sense, realizations of event sequences. 
It does not depend on assumptions about the probability space. 
For this realization, we need to create a subset of similar event sequences.
For this, we cluster from a pre-selected set of realizations, a training sample.
We use this method as it is the most general one.

\subsection{Finding the optimal forecast}
We solve the optimization problem in \autoref{eq1} by replacing $\mathcal{P}\left(\tau\leq\theta\mid {\mathcal{F}}_t\right)$ and $\mathsf{E} \left(G\mid {\mathcal{F}}_t\right)$ with estimates based on the realizations of close event sequences. 
We describe the algorithm for finding close realizations of event sequences in \autoref{seq: clustering}.
Let us have a sequence of events $p = (a_1,\ldots,a_k,\ldots)$ and the training sample $G$. 
We want to predict when the next $a_{k+1}$ event appears in this sequence.
We denote with $N_{(t, T)}$ the number of events in a given sequence on a time interval $(t, T)$, where $t$ is the current time.
With this, it is possible to construct the following function

\begin{equation*}
    f(k,t,T) = P(N_{(t,T)} = k | \mathcal{F}_t) \approx \frac{\mathcal{E}_m^{k_{(t,T)}}}{\mathcal{P}_m}
\end{equation*}

where $\mathcal{E}_c^{k_{(t,T)}}$ represents the number of event sequences from cluster $m$ with $k$ events on the segment $(t,T)$. 
The number of sequences in the cluster $m$ is $\mathcal{P}_m$, thus $k = 0,1,\ldots, M(t,T)$.
Since, obviously, there is an integer $M(t,T)\ge 0$ such that $f(M(t,T)+i,t,T)=0$ for $i\in \mathbb{N}$.
In this case, ${\mathcal{F}}$  indicates the information that we "capture" from the sequences in the cluster.
We look for the time $\tau^*\in (t,\infty)$.
It is the moment in time achieving a minimum as following

\begin{equation}\label{eq4}
    S(\tau,c)=P\left(\tau\leq\theta\mid {\mathcal{F}}_t\right) + c\mathsf{E} \left(G\left([\tau-\theta]^+\right)\mid {\mathcal{F}}_t\right),
\end{equation}

To simplify the notation, we omit $\mathcal{F}_t$ in the next steps. 
We notice that
$\mera\ls\tau\leq\theta\rs = \mera\ls N_{(t,\tau)}=0\rs =
1-\mera\ls N_{(t,\tau)}\geq 1\rs$.
Hence, minimizing the first term in the criterion is equivalent to maximizing the probability that on the interval $(t,\tau)$ at least one event appears. 
The second term plays the role of regularization.
It controls  that the value of $\tau$ is not too large, and the event is not "obsolete."
It is obvious that $\mera\ls\theta<\tau\rs = 1-\mera\ls\theta\geq\tau\rs$ and this equals $1-\mera\ls N_{(t,\tau)}=0\rs = 1-f\ls0,t,\tau\rs$ where

\begin{equation*}
    f\ls0,t,\tau\rs = \frac{\mathcal{D}^{k_{(t,\tau)}}_m}{\mathcal{Q}_m}
\end{equation*}

The number of sequences that have no events on $(t, \tau)$ is $\mathcal{D}^{k_{(t,\tau)}}_m$ and $\mathcal{Q}_m$ is the total number of sequences in the cluster.
There are $n \leq \mathcal{Q}_m$ time periods $t = t_0 < t_1 < \ldots < t_n$.
Similarly, we have numbers $1 = \alpha_0 > \alpha_1 > \ldots > \alpha_n \geq 0$ such that $f(0,t,\tau) \equiv \alpha_i$ when $\tau \in [t_i,t_{i+1}), i = 0,\ldots,n$.
In this case, it is $t_{n+1} = \infty$. 
Therefore, $\alpha_i = \frac{\mathcal{D}^{k_{[t_i, t_{i+1})}}_m}{\mathcal{Q}_m}$ represents the number of sequences that have no events on $[t_i,t_{i+1})$ divided by the number of sequences in the cluster.
Thus, we have $\mera\ls\theta<\tau\rs = 1-\alpha_i$ for $\tau\in[t_i,t_{i+1})$. 
We denote the distribution function of the random variable $\theta$ as $F(s)=\mera\ls\theta<s\rs$.
As a next step, we want to compute $\me G\ls[\tau-\theta]^+\rs$.
We assume that $\tau\in[t_i,t_{i+1})$, $i=0,\ldots,n$. 
With this, we have

\begin{align*}
\me G\ls[\tau-\theta]^+\rs& = \int_t^{\infty}G\ls\max[\tau-y,0]\rs dF(y)\\
&=\int_t^{t_i}G\ls\tau-y\rs dF(y)+\int_{t_i}^{\tau}G\ls\tau-y\rs dF(y)\\
&=\int_{t}^{t_i}\ls\tau-y\rs dF(y).
\end{align*}

\paragraph{Linear penalty function}
We consider the case when $G(x) = x$. 
Then, when $\tau\in[t_i,t_{i+1})$, $i=0,\ldots,n$, we have

\begin{align*}
\me \ls\tau-\theta\rs^+& = \int_{t}^{t_i}\ls\tau-y\rs dF(y)\\
&=\tau F(t_i) - \sum_{j=0}^{i-1}t_{j+1}
dF(t_{j+1})=\tau\ls1-\alpha_i\rs \\
&-\sum_{j=0}^{i-1}t_{j+1}\ls\alpha_j-\alpha_{j+1}\rs.
\end{align*}

Therefore, we need to find $\tau^*\geq t$, such that the expression in \autoref{eq1} takes the minimum value, and for $\tau\in[t_i,t_{i+1})$, $i=0,\ldots,n$ to have the form

\begin{equation}\label{lineq2}
    S(\tau,c)=\alpha_i+c\tau\ls1-\alpha_i\rs-c\sum_{j=0}^{i-1}t_{j+1}\ls\alpha_j-\alpha_{j+1}\rs.
\end{equation}

From \autoref{lineq2}, we know that \autoref{eq1} takes the minimum value for $\tau\in\{t_1,\ldots,t_n\}$. 
Thus, to find  $\tau^*$, we need to find an optimal $j^*$, such that  $S(t_{j^*},c)$ takes the minimum value among the values $S(t_1,c),S(t_2,c),\ldots,S(t_n,c)$, then $\tau^*=t_{j^*}$.

\subsubsection{Nonlinear penalty function}
Let us consider the case when $G(x)$ is a nonlinear function.
We require that $\underset{x\to+\infty}{\lim} G(x) = \mathrm{const}<\infty$. 
For this, we can use the following types of the $G(x)$ function, where we assume that $\beta>0$ is a parameter, namely:

\begin{equation}\label{eee1}
    G(x) = \frac{e^{\beta x}-e^{-\beta x}}{e^{\beta x}+e^{-\beta x}} ~\textbf{or}~ G(x) = \frac{x^{\beta}}{1+x^{\beta}}
\end{equation}

For both cases, we have that $\underset{x\to+\infty}{\lim} G(x) = 1$.
In this case, when $\tau\in[t_i,t_{i+1})$, $i=0,\ldots,n$, we have

\begin{align*}
\me G\ls[\tau-\theta]^+\rs& = \int_{t}^{t_i}G\ls\tau-y\rs dF(y)\\
&=\sum_{j=0}^{i-1}G\ls\tau-t_{j+1}\rs\ls\alpha_j-\alpha_{j+1}\rs.
\end{align*}

Therefore, we need to find $\tau^*\geq t$, such that \autoref{eq1} takes the minimum value and for $\tau\in[t_i,t_{i+1})$, $i=0,\ldots,n$ to have the form

\begin{equation}\label{nelineq2}
    S(\tau,c)=\alpha_i+c\sum_{j=0}^{i-1}G\ls\tau-t_{j+1}\rs\ls\alpha_j-\alpha_{j+1}\rs.
\end{equation}

From \autoref{nelineq2}, we observe that \autoref{eq1} takes the minimum value when $\tau\in\{t_1,\ldots,t_n,\infty\}$. 
So, to find $\tau^*$, we need to find such an optimal  $j^*$, such that $S(t_{j^*},c)$ takes the minimum value among the values \linebreak 
$S(t_1,c),S(t_2,c),\ldots,S(t_n,c)$.
For this, we set $\tau^*=t_{j^*}$ if  $S(t_{j^*},c)<c$.
Otherwise, we have $\tau^* = \infty$.
Of course, the criterion in \autoref{eq1} depends on the parameter $c$.
We still needs to define it.
At first glance, we might assume that it is easier to get the optimal $\tau^*$.
In this case, we only need to minimize the criterion

\begin{equation}\label{remeq3}
    \me G\ls|\tau-\theta|\rs.
\end{equation}

Further, we do not need to select $c$ from the data.
However, this is not the case.
Whereas, for the criterion in \autoref{eq1}, we can find the exact value of the minimum in linear time.
We can do this by iterating over the values of $\tau\in\{t_1,\ldots,t_n,\infty\}$.
For the criterion in \autoref{remeq3}, we can reach the minimum value when $\tau$ is not in the $\{t_1,\ldots,t_n,\infty\}$ values.

\subsubsection{Conditional criteria}
We note that $j^*=j^*(c)$. 
Therefore, we can show that in the case of a linear penalty function, for any $j\in\{1,2,\ldots,n\}$, there is $c^*$.
In this case, we have $j^*(c^*)=j$.
Further, we can take an arbitrary $k\in\{1,\ldots,n\}$. 
By definition, for $t_{k}\in\{t_1,\ldots,t_n\}$, we have the equality  $\mera\ls t_k\leq\theta\rs=\alpha_k$. 
Next, we select $c^*$ such that $j^*(c^*)=k$
Similarly, we choose the value of $\tau$ as $\tau^*=t_{j^*(c^*)}=t_k$.
It is the minimum value of \autoref{eq2}.
We reach it when we select $c=c^*$.
Therefore, we have

\begin{equation*}
    \underset{\tau\geq
    t}{\min}\ls\mera\ls\tau\leq\theta\rs+c\me\ls\tau-\theta\rs^+\rs.
\end{equation*}

It follows that $\tau^*=\tau^*(c^*)$ for the selected $c^*$ is the solution of the conditional optimization problem

\begin{equation*}
    \underset{\tau\geq t}{\min}\me\ls\tau-\theta\rs^+
\end{equation*}

provided that $\mera\ls\tau\leq\theta\rs\leq\alpha_k$.

\subsection{An engineering approach to forecasting}
Our proposed mathematical model is computationally intensive.
We propose a slightly different approach for determining the optimal time point, $\tau^*$.
The essence of the method is as follows.
We start by calculating the values of $\tau^*$ for each cluster in advance when we create the cluster. 
We let the cluster size be $Q$.
Thus, $ W $ sequences $(W\le Q)$ from the cluster have events after time $T_i$. 
Similarly, $Q-W$ sequences from the cluster have no events after time $T_i$.
We denote the periods when each of the sequences has the first events after the time $T_i$ as $t^1,\ldots,t^W$. 
We define one value for each sequence.
Further, we order these times in ascending order.
As a result, we get $t^{(1)},\ldots,t^{(W)}$. 
We let t$\alpha\in(0,1)$ be the index $D~=~[\alpha Q]$ for the given parameter.
Therefore, we define the value of $tau^*$  as following

\begin{equation}\label{eng1}
\tau^* =
\left\{
\begin{array}{l}
t^{(D)}, \mbox{ if} D\le W
\vspace{1pt}
\\
\infty, \mbox{ if} D>W
\end{array}
\right.
\end{equation}
 
If we have $t^*=\infty$, then we do not register any more events for this sequence.
Let us look at a specific example.
\autoref{eng1} implies that if $\alpha = 0.2$ for some value, $T_i\in[0,T], i=1,\ldots,K$, in the cluster after the time $T_i$, then less than 20\% of sequences have events.
Thus, we have $t^*=\infty$, and we consider that the given sequence is also unlikely to have new events.

\section{COHORTNEY}\label{seq: clustering}
We need to learn how to find a set of instances.
They must be close, in some sense, to a given realization of an event sequence.
Our objective is to evaluate two expressions.
We want to obtain $\mathcal{P}\left(\tau\leq\theta\mid {\mathcal{F}}_t\right)$.
Similarly, we want $\mathsf{E} \left(G\mid {\mathcal{F}}_t\right)$.
Therefore, we must split the training sample into clusters.
For a new realization of the event sequence, we must find the cluster closest to it in order to estimate the probability and expectation.
Unlike other algorithms, COHORTNEY does not need to specify any initial approximations, the number of clusters, or the amount of connectivity.
The input is only a set of objects with a specified proximity function.
With this function, for any objects $x_i$ and $x_j$, it is possible to determine the distance $\rho_{ij} = \rho(x_i, x_j)$ between them.
Therefore, we must introduce a measure of proximity between instances of event sequences.
Let us assume that we have the numerical characteristics of two instances of event sequences, $(a_1, a_2,\ldots, a_n)$ and $(b_1, b_2,\ldots, b_m)$
In general, $n\neq m$, so the usual Euclidean distance is not suitable here.
In this case, the DTW distance function can be an option.
However, there are two disadvantages to using them.
First, it has high computational complexity.
It is $O(nm)$.
Second, DTW considers two events as one, if the event sequence in question has several events that are close in time.
Therefore, it considers the sequence to be similar to the one that has only one event at the same time interval.
As a result, the measure of proximity between instances is small.
We can solve the first difficulty using a lower bound.
We cannot address the second limitation.
Therefore, we need a measure of proximity devoid of these disadvantages.

\subsubsection{Measure of proximity between event sequences}
We consider an arbitrary event sequence $p = (a_1, a_2 \ldots a_n)$, $a_i \le a_{i+1}$.
Here, $a_i$ is the time when the $i$-th event occurs.
Let there be some parameter $t$, which we call the present moment.
$t$ is fixed and satisfies the inequality $a_n\le t$.
We introduce a weight function that shows the density of the event sequence in a specific time interval.
For this, we consider an arbitrary time interval $(t_1, t_2)$.
Then the weight function that characterizes the number of events that occurred in this time interval is equal to

\begin{equation*}
fws(p, t_1, t_2) = \min\left([{\log_2{\left(n + 1\right)}}],9\right)\mbox{.}    
\end{equation*}

We have $n$ as the number of events in the time interval $(t_1, t_2)$, $n = 0, 1, \ldots$.
In this case, $[\cdot]$ is the integer part of the number.
Now we define a weight function for the entire event sequence. 
We consider an arbitrary partition $\Delta T^*$  of the half-interval $(0;t]$.
The function has the form,

\begin{equation}\label{mh_razb}
\Delta T^* = \{t_0, t_1 \ldots t_m\} \mbox{ , } 0 = t_0 \le t_1 \le \ldots \le t_m = t.
\end{equation}

Let us calculate the total value of the interval weights for this partition and the time $t$.
We compute $fw(p, t, \Delta T^*)$ as following

\begin{equation*}
fw(p, t, \Delta T^*) = \{fws(p, t_0, t_1), \ldots , fws(p, t_{m-1}, t_m)\}.    
\end{equation*}

Thus, we can characterize the sequence of events using a triplet of values.
They are $\{t, \Delta T^*, fw(p, t,\Delta  T^*)\}$.
This triplet is the weight function for the entire event sequence.
Let us now consider the uniform partition, $\Delta T^n$,  of the half-interval $(0;t]$.
For example, we have $t_i - t_{i-1} = \Delta t_n = \left[\frac{T}{2^n}\right]$ in \autoref{mh_razb}.
We call $\Delta t_n$ the partition scale. 
Subsequently, we characterize the sequence of events by the triplet, $\{t, {\Delta t}_n, fw(p, t, \Delta T^n)\}$.
We now increase the value of  $n = 0, 1, \ldots$.
Thereby, we reduce the value of ${\Delta t}_n$ until $\left[\frac{T}{2^n}\right] \ge \delta$.
Here, $\delta$ is a pre-set value.
It sets the minimum distance between events in the event sequence.
Let there be two sequences of events. 
If for some fixed $n$, their triplets coincide, then we call these sequences coincident on the given partition scale.
Accordingly, the smaller the difference between these three, the closer these two sequences are from each other on the split scale.
We build an algorithm based on the proximity measure.

\subsubsection{The clustering algorithm}
We have a numerical sample of events, $(a_1, a_2, \ldots, a_n)$.
We want to cluster them.
The data has two characteristics.
First, it is $a_1\le a_2\le\ldots\le a_n$.
Second, different samples generally have different lengths.
There are also sequences without events. 

\paragraph{Source data}
The source data is a selection of known instances of event sequences $P = (p_1, p_2, ... p_M)$, with the above properties.
We call this sample a training sample.
Using this sample, we create groups of "similar" event sequences as follows.

\paragraph{Output}
The output of the algorithm is a set of clusters.
We describe each cluster by a triplet of values $\{t, \Delta T^n, fw(p, t, \Delta T^n)\}$.
Therefore, the cluster consists of such instances of event sequences where these triplets coincide.

\paragraph{Initialization}
We set the values $T_h$, $T_b$, $\gamma$, $N$.
They are parameters of the clustering algorithm.
$T_h$ is the time threshold value. 
We assume that no events occur when $t > T_h$.
We define $T_b$ as a time value.
Further, we have $\gamma > 1$ as a scaling parameter.
Here, $T_b$ and $\gamma$ set the grid parameters.
Our minimum cluster size is $N$.
We take the time grid $T^*$ with nodes $T_j = {\gamma}^jT_b$, $j\in J\subset\mathbb{Z}$ on the interval $(0, T_h]$.

\paragraph{Iteration of the algorithm}
We now consider the $i$-th iteration of the algorithm $(i = 1 \ ldots K)$.
On the $i$-th iteration of the algorithm, we assume that $t = T_j$.
In this case, COHORTNEY performs the following steps.

\begin{description}
  \item[Step 1.] 
  COHORTNEY takes an arbitrary node $T_j$ from the grid.
  \item[Step 2.] 
  For each $n$ and for each $p_k\in P$, it calculates the weight function.
  \item[Step 3.] 
  For each $n$, it groups those $p_k$ that have the same triplets $(T_j, {\Delta T}^n_i, fw(p_k, T_j, {\Delta T}^n_i))$.
  They will form a cluster.
  It uses this triplet as identifiers of the cluster.
  Then, it groups each $p_k$ at each step of the iteration.
  In other words, each iteration creates additional clusters.
  The same event sequence $p_k$ can be in different clusters for different $n$.
  \item[Step 4.] 
  It discards those clusters that are smaller than the minimum size, $N$.
\end{description}

\textbf{Example.} 
For an arbitrary $T_i$ in the first step, $(n = 0)$, COHORTNEY generates the following clusters,

\begin{equation*}
    (T_i, T_i, 0), (T_i, T_i, 1), (T_i, T_i, 2), \ldots
\end{equation*}

In the second step, where $(n = 1)$, COHORTNEY has

\begin{equation*}
\left(T_i, \frac{T_i}{2^1}, 01\right), \left(T_i, \frac{T_i}{2^1}, 11\right), \left(T_i, \frac{T_i}{2^1}, 20\right), \ldots
\end{equation*}

COHORTNEY generates for the third step, when $(n = 2 )$, 

\begin{equation*}
\left(T_i, \frac{T_i}{2^2}, 0100\right), \left(T_i, \frac{T_i}{2^2}, 1411\right), \left(T_i, \frac{T_i}{2^2}, 0220\right), \ldots
\end{equation*}

COHORTNEY applies this clustering procedure to all grid nodes.
The result is a set of clusters.
They correspond to all grid nodes and all possible partitioning scales.
After the clustering, COHORTNEY calculates the optimal $\tau^*$ for each cluster, respectively for each $T_j$, in advance.

\subsubsection{Search for the nearest cluster}
With the arrival of new event sequences, the problem arises of finding the most similar sequences from the training sample.
How can we find the nearest cluster?
Let us describe the algorithm for finding the nearest group.

\paragraph{Input}
The input is a set of clusters that COHORTNEY builds based on the training sample.
For training, COHORTNEY uses a new realization of the event sequence $p_i = (a_{j1}, a_{j2}, \ldots , a_{jk}, \ldots)$ and the current time $t$.
It assumes that all $a_{j_k}\le t$.

\paragraph{Output}
The output is $ \tau^*$.
It is the optimal time for the event to occur.

\paragraph{Algorithm description}
\begin{description}
  \item[Step 1.] 
  COHORTNEY finds in the grid $T^*$ the nearest$ T_t $ node to $t$ that does not exceed $t$.
  
  \item[Step 2.] 
  Cohortney iterates over the value $n = 0 \ ldots K$. 
  For any $n$, COHORTNEY builds a partition $\Delta T^n$ on the segment $[0, T_t]$.
  It then calculates the weight function $w_n = fw(p_i, \Delta T^n, T_t)$.
  
  \item[Step 3.] 
  COHORTNEY looks for a cluster corresponding to a specific ID.
  In this case, we have $I_n = (T_t, \Delta T^n, w)$.
    \begin{enumerate}
    \item First possibility, if $n = 0$ and there is no exact match $I_0$, then COHORTNEY chooses the cluster $I_f = (T_t, \Delta T^n, f)$ from the set of clusters, where $f$ is closest to $w_n$.
    In other words, COHORTNEY selects the closest cluster to the given realization of the event sequence at the specified partitioning scale.
    \item Second possibility, if COHORTNEY finds the cluster, then $n = n + 1$ and COHORTNEY goes back to step 1.
    \item Third possibility, COHORTNEY does not find a cluster. 
Hence, it takes $I_{n-1}$ as the closest cluster.
    \end{enumerate}
\end{description}

In other words, in any case, during the first step, COHORTNEY finds the nearest cluster, even if it cannot find an exact match for the ID.

\subsection{General structure of the prediction algorithm}

\textbf{Input:}
    \begin{enumerate}
    \item The current time $t$.
    \item The time points $0\le t_1,\ldots,t_N\le t$ when events occur for a given event sequence realization.
    \item The parameters of the algorithm.
    \end{enumerate}

\textbf{Output:}
   \begin{enumerate}
    \item The optimal time $\tau^* > t$  when a new event appears in the event sequence.
    \end{enumerate}


\section{Problem statement and method of its solution}
The task is as follows.
We have a post,$p = (a_1, a_2, \ldots, a_k)$, and its comments.
We define $a_i$ as the times when comments appear in the post, $p$.
With this data, we predict a moment in time, $\tau^*$, when the next comment, $a_{k+1}$, appears in the time interval, $(a_k, \tau^*)$.
To do this, COHORTNEY takes a training sample of posts. 
It then performs clustering. 
After this, it calculates the optimal time, $\tau^*$, for each cluster.
Next, for each newly received post $p_i$, COHORTNEY looks for the nearest cluster and estimates the optimal time for the comment to appear.

\subsubsection{Description of the data structure}
The object of research is a post in social media and all known comments related to this post.
We only know the publication times of the post and comments.
They are known as the source data: $p_i^* = (t_0^*, t_1^*,\ldots,t_n^*)$.
Here, $t_0^*$  is the publication time of the post.
Moreover, $t_i^*$ is the creation time for the comment $i$.
We express all values in seconds.
Let us take the time of publication of the post as the starting point. 
We transform the publication time of the comments into the form

\begin{equation*}\label{data1}
    p^* = (t_0^*, {\Delta t}_1, {\Delta t}_2, \ldots, {\Delta t}_n)\mbox{.}
\end{equation*}

In this case, ${\Delta t}_i = (t_i^* - t_0^*) > 0$, as comments only appear after the publication date of the post and ${\Delta t}_i \le {\Delta t}_{i+1}$.
We note that ${\Delta t}_n$ is the time of the last known comment up to the current time, i.e. ${\Delta t}_n < t$, where $t$ is the current time.

\section{Modeling}

\paragraph{Description of the training test sample}
As a training sample, we use a set of 83,000 posts with comments.
We select them randomly and transform them into the form in \autoref{data1}.
We select the posts from historical data.
As a result, new comments for these posts should not appear.
In total, there are around 373,000 comments on posts in this sample.
Approximately half of all posts do not have comments.

\section{Experiments}
This work proposes to use two different types of clustering algorithms with COHORTNEY. 
We call them COHORT1 and COHORT2.

\paragraph{Deterministic Approach}
For a given post and the current time $t_r$, we look through the hours ranging from 1, 3, 12, 24, 168 to 744.
Each time we check whether at least one comment appears.
If a comment does not appear after the last step, we assume that there are no more comments in this post.

\paragraph{COHORT2}
This algorithm is based on clustering and calculating the optimal comment appearance time for each cluster, as we describe above.
For the current moment $ t = t_r $, we find the cluster from the training sample closest to the given post. 
We take the $ \tau^* $ calculated for the selected cluster and at the time $ \tau^* $ we check if the post has a comment.
If a comment does not appear, then we set the current time to $ t = \tau^* $.
We define a set of posts from the training sample, similar to the given post, taking into account the fact that now the current time is $ t = \tau^* $.
We calculate the new $ \tau^* $ and at time $ \tau^* $ check if a comment has appeared, and iterate. 
Let us suppose that at some step it turned out that $ \tau^* = \infty $, or less than $ \alpha \cdot 100\% $ posts from the resulting cluster contain comments. 
In that case, we do not continue analyzing this post for comments.
The algorithm stops for this post, and it transitions to analyze another post from the test sample.

\paragraph{Evaluation}
Let $ \theta_ {i, 1}, \ldots, \theta_ {i, k_i} $ be the posting times on the time interval $ (\tau_{i-1}^*, \tau_{i}^*] $, where
$ \tau_i^* $ is the $i$-th moment of peeping.
We obtain it from using one of our algorithms.
The initial time is $ \tau_{0}^* = 0 $. 
We denote the number of posts in the training set as $M$.
Further, $N$ is the number of peeks that we do for a post until we stop.
It is the time after which the algorithm no longer evaluates this post.
For example, it will not turn out that $\tau_{N+1}^*=\infty$ or there are few posts with comments in the corresponding cluster.

We evaluate the quality of algorithms as follows.

\begin{enumerate}
    \item \textbf{Time delay for the post.} 
    First, we define the vector $\Delta = \ls \Delta_1,\ldots,\Delta_M\rs$.
    It has the components 
    \begin{equation*}
        \Delta_i = \sum_{i=1}^{N}{\left(\sum_{j=1}^{k_i}\ls\tau_{i}^*-\theta_{i,j}\rs\right)}.
    \end{equation*}
    
    Let's calculate the average, median, and $95\%$-quantile of these values from the test sample, respectively, and build the dependence of the resulting characteristics on $\alpha$.
    This value characterizes the total time delay per post when determining the optimal time for the comment to appear.
    Fig.~\autoref{post-penalty-mean}, fig.~\autoref{post-penalty-median}, fig.~\autoref{post-penalty-q95}.

    \item \textbf{Time delay for a comment.} 
    Define the vector \linebreak 
    $\delta=\ls \delta_1,\ldots,\delta_M\rs$.
    
    Its components $\delta_i$ equals $\sum_{i=1}^{N}{\left(\frac{1}{k_i}\sum_{j=1}^{k_i}\ls\tau_{i}^*-\theta_{i,j}\rs\right)}$. 
    Let's calculate the average, median, and $95\%$-quantile of these values from the test sample, respectively, and build the dependence of the resulting characteristics on $\alpha$.\\
    This value characterizes the average time delay per comment in a post when determining the optimal time for a comment to appear.
    Fig.~\autoref{comment-penalty-mean}, fig.~\autoref{comment-penalty-median}, fig.~\autoref{comment-penalty-q95}.

    \item \textbf{Relative intensity per post.} 
    Define the intensity as $\lambda=\frac{1}{N}\sum_{i=1}^{N}\frac{1}{\tau_i^*-\tau_{i-1}^*}$.
    Using the test sample, we calculate the average value of $\bar{\lambda}$, the median of $m\ls\lambda\rs$, $95\%$ - the quantile of $q\ls\lambda\rs$ and build the dependence of these values on $\alpha$. 
    Fig..~\autoref{relative-intensity-mean}, fig.~\autoref{relative-intensity-median}, fig.~\autoref{relative-intensity-q95}.

    \item \textbf{Probability.} 
    Let's estimate the probability of finding a comment using the formula
    $p=\frac{1}{N}\sum_{i=1}^N I$ in $(\tau_{i-1}^*,\tau_{i}^*]$ if there are comments, where $I(A)$ is an event indicator $A$.
    Using the test sample, we calculate the average value of $\bar{p}$, the median of $m\ls p\rs$, and $95\%$ - the quantile of $q\ls p\rs$ and build the dependence of these values on $\alpha$.
    Figure~\autoref{probability-mean}, figure~\autoref{probability-median}, figure~\autoref{probability-q95}.

    \item \textbf{Absolute intensity.} 
    For each post from the test sample, we calculate the value $N$ - the number of views.
    Using the test sample, we estimate the average value of $\bar{N}$, the median $m\ls N\rs$, and $95\%$- the quantile $q\ls N\rs$ and construct the dependence of these values on $\alpha$.

    Fig.~\autoref{absolute-intensity-mean}, fig.~\autoref{absolute-intensity-median}, fig.~\autoref{absolute-intensity-q95}.
\end{enumerate}

Based on the results of the simulation, the value $\alpha$ was chosen so that with the same intensity of work, the time delay in predicting the time of comment appearance was minimal.

\subsection{Results of the actual operation of the algorithm}
\begin{figure}[ht]
\centering
\includegraphics[width=\columnwidth]{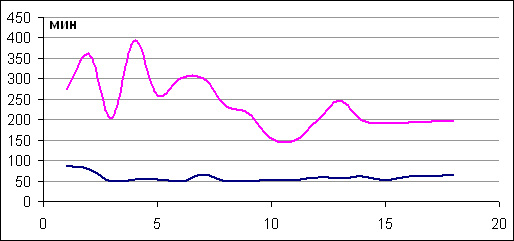}
\caption{\textbf{Comparison of delay medians for the COHORT2(blue) and production(pink) algorithms}}
\label{mega-hash_production}
\end{figure}

The proposed algorithm was implemented in the system.
Every 20th post is processed by the algorithm.
The table shows the median delay in downloading comments for the production and mega-hash Cohortney algorithms.
In figure~\autoref{mega-hash_production} It can be seen that the median delay for the production algorithm experiences stronger fluctuations than for the mega-hash algorithm.

\begin{center}
\begin{tabular}{|l|c|c|c|}
\hline
\textbf{Period} & \textbf{mega-hash,} & \textbf{production,} & \textbf{Relationship}\\
\textbf{Time} & \textbf{min} & \textbf{min} & \textbf{delay}\\
\hline
1 & 87 & 273 & 3,14\\
\hline
2 & 79 & 359 & 4,54\\
\hline
3 & 49 & 204 & 4,16\\
\hline
4 & 53 & 394 & 7,43\\
\hline
5 & 53 & 259 & 4,89\\
\hline
6 & 50 & 300 & 6,00\\
\hline
7 & 65 & 301 & 4,63\\
\hline
8 & 50 & 234 & 4,68\\
\hline
9 & 50 & 216 & 4,32\\
\hline
10 & 51 & 155 & 3,04\\
\hline
11 & 51 & 150 & 2,94\\
\hline
12 & 58 & 197 & 3,40\\
\hline
13 & 56 & 246 & 4,39\\
\hline
14 & 62 & 198 & 3,19\\
\hline
15 & 52 & 191 & 3,67\\
\hline
16 & 61 & 192 & 3,15\\
\hline
17 & 62 & 196 & 3,16\\
\hline
\end{tabular}
\end{center}

\section{Conclusion}
On real data, the delay in downloading comments for the proposed algorithm is 4-6 times less.
You need to check that for the specified value $\alpha\approx 0.2$, the COHORT2 and production intensities really match. Perhaps the value of $\alpha\in(0,1)$ must be adaptively adjusted in some way so that the intensities coincide (if it turns out that over time the intensity values begin to differ greatly from each other).
You need to build clusters based on a larger training sample, and the value of the $T_h$ parameter should be even larger, and the $T^*$ grid even thicker.
This allows to more accurately determine which cluster a given post belongs to.
It is necessary to introduce a different concept of time-to measure time in the number of comments written so far.
This allows you to account for variations in the number of comments associated with daily, weekly, and monthly cycles.

\section{Application}
\subsection{DTW proximity function}
The DWT (Dynamic Time Warping) function takes as input two sequences of arbitrary length and considers all options for mapping one sequence to another.
As a measure of comparison of these sequences, the minimum "cost" or minimum "path" of transferring one sequence to another is taken.
Let's consider this at a formal level.

Suppose we have two sequences of events $Q$ and $C$ of length $n$ and $m$, respectively

\begin{equation*}
    Q = q_1, q_2,\ldots,q_i,\ldots,q_n
\end{equation*}

\begin{equation*}
    C = c_1, c_2,\ldots,c_j,\ldots,c_m
\end{equation*}

To compare these two sequences, create a matrix $ nxm $ where $ (i, j) $ is the matrix element - the distance $ d (q_i, c_j) $ between the points $ q_i $ and $ c_j $, i.e. $d(q_i, c_j) = {\left(q_i - c_j\right)}^2$.
Each element of the matrix reflects the level of closeness of two events in different sequences.
A path $W$ (\emph{warping path})) is a set of adjacent elements that define the mapping of $ Q $ to $ C $.
Denoting the k-th element of $ W $ as $w_k = {(i,j)}_k$, we have:

\begin{equation*}
    W = w_1, w_2, \ldots, w_k, \ldots, w_K\mbox{,  } max(m,n)\leq K < m+n-1
\end{equation*}

This path satisfies several restrictions:

\begin{enumerate}
\item \textbf{Boundary condition.} 
$w_1 = (1,1)$ and $w_k = (m,n)$. 
This corresponds to the fact that the path starts and ends on diagonally opposite elements of the matrix.

\item \textbf{Continuity.} 
Let be $w_k = (a,b)$ and $w_{k-1} = (a', b')$, then $a-a'\le 1$ and $b-b'\le 1$. 
In other words, adjacent elements in the path $W$ can only be adjacent elements of the matrix (including the diagonal)

\item \textbf{Monotony.} 
Let be $w_k = (a,b)$ and $w_{k-1} = (a', b')$, then $a-a'\ge 0$ and $b-b'\ge 0$. 
That is, the movement along the path should occur with each new step.
\end{enumerate}

Obviously, there are many paths that satisfy these constraints, but we are only interested in the path that minimizes the functional:

\begin{equation*}
  DWT(Q,C) = min\left(\sqrt{\sum_{k=1}^K{w_k}}\right)  
\end{equation*}

This path can be found using dynamic programming to calculate the following recurrence relation, which is defined by the cumulative distance $\gamma(i,j)$  as the distance $d(i, j)$ in the current cell and the minimum cumulative distance to adjacent cells:

\begin{equation*}
\gamma(i,j) = d(q_i,c_j) + min\{\gamma(i-1,j-1), \gamma(i-1,j), \gamma(i,j-1)\}
\end{equation*}

The Euclidean distance between two sequences can be obtained as a special case of DTW, where the k-th element of the path is $W$ $w_k = {(i,j)}_k$~, $i=j=k$.
Note that this corresponds to the special case when two sequences are of the same length.
In General, the DTW count is $O(nm)$.

\subsection{Score for DTW}
Consider the lower bound for the DTW function, which is computed faster than the DTW itself.
Let there be given two sequences $\textbf{x} = (x_1, \ldots, x_m)$  and $\textbf{y} = (y_1, \ldots, y_n)$, denote by $max(\textbf{x})$  and $max(\textbf{y})$  are the maximum elements in the sequences $max(\textbf{x})$ and $max(\textbf{y})$, respectively, and after $\textbf{x}$ and $\textbf{y}$ - minimum elements.
Without loss of generality, we  assume that $max(\textbf{x}) \ge max(\textbf{y})$.
The method described below is based on the following statement:

\begin{equation*}
   \mid max(\textbf{x})-max(\textbf{y})\mid\le D_{warp}(\textbf{x},\textbf{y}) 
\end{equation*}

where $D_{warp}(\textbf{x},\textbf{y})$ is the distance between $\textbf{x}$ and $\textbf{y}$ calculated using the DTW function.
Let be $R_x = \left(min(\textbf{x}), max(\textbf{x})\right)$, $R_y = \left(min(\textbf{y}), max(\textbf{y})\right)$, the following mutual positions of these intervals are possible:

\begin{enumerate}
\item $R_x$ and $R_y$ intersect $\left(min(\textbf{x})\le max(\textbf{y}), min(\textbf{x})\ge min(\textbf{y})\right)$
\item $R_x$ includes $R_y$ $\left(min(\textbf{x}) < min(\textbf{y})\right)$
\item $R_x$ and $R_y$ do not intersect $\left(min(\textbf{x}) > max(\textbf{x})\right)$
\end{enumerate}

Define the distance function $D_{lb} ()$ as follows:

\begin{align*}
D_{lb}(\textbf{x},\textbf{y}) = \left\{
\begin{array}{l}
\sum_{x_i > max(\textbf{y})}{\mid x_i - max(\textbf{y})\mid} \\
+ \sum_{y_j < min(\textbf{x})}{\mid y_j - min(\textbf{x})\mid} \mbox{if case 1}
\vspace{3pt}
\\
\sum_{x_i > max(\textbf{y})}{\mid x_i - max(\textbf{y})\mid} \\
+ \sum_{x_i < min(\textbf{y})}{\mid x_i - min(\textbf{y})\mid} \mbox{if case 2}
\vspace{3pt}
\\
max( \sum_{i=1}^m{\mid x_i - max(\textbf{y})\mid}, 
\\\sum_{i=1}^n{\mid y_j - min(\textbf{x})\mid} ) \mbox{if case 3}
\end{array}
\right.
\end{align*}

It is obvious that $D_{lb}$ can be calculated in time $O(n+m)$ and for any sequences $\textbf{x}$ and $\textbf{y}$:

\begin{equation*}
    D_{lb}(\textbf{x},\textbf{y})\le D_{warp}(\textbf{x},\textbf{y})
\end{equation*}

\subsection{Spectrum clustering algorithm}
Consider a set of elements  $p_1, \ldots, p_M$  of arbitrary nature.
It is necessary that for each pair of elements $p_i$ and $p_j$, a number $\alpha_{ij}$ is set, which can characterize the degree of" proximity " between these elements.
We assume that the higher this number, the stronger the relationship between the corresponding elements.
Thus, we can assume that the considered set of elements $p_1, \ ldots, p_M$ is characterized by the matrix $A = {\left\{\alpha_{ij}\right\}}_{i,j=1}^M$.
We call it the connection matrix.
The purpose of the processing matrix is the partition of the whole set of elements such disjoint subsets $G_1, \ldots, G_L$, we call them clusters that value when $\alpha_{ij}$ between the elements in one subset, if possible, large, and between elements in different subsets -- little.
This problem is called the clustering problem of the original set of elements.
The idea of the existence of sufficiently isolated clusters of strongly connected elements means that we are referring to two, although similar in content, but nevertheless different properties of such aggregates.
On the one hand, the clusters in question are characterized by larger (in a certain average sense) values of the relationship between their elements than other clusters.
On the other hand, isolated clusters consisting of strongly connected elements can be characterized by the following property.
The "nearest'' an element is to any subset of elements from one cluster, the likeliest it is that this element belongs to the cluster. 
Any element from another cluster is significantly more "distant" from this subset of elements.
We use this property below.
We introduce the proximity measure $K (G, p_l)$ between a set $G$ containing a certain number of$ m $ elements and a single element $p_l$:

\begin{equation}\label{eq2}
K(G,p_l) = \frac 1 m\sum_{i\in G}{\alpha_{il}}
\end{equation}

The algorithm uses the original set of elements $p_1, \ldots, p_M$ to construct their ordered sequence$p_{l_1},p_{l_2},\ldots,p_{l_M}$.
An arbitrary element is selected as $p_{l_1}$.
Then, from the remaining elements, the element closest in terms of magnitude (\autoref{eq2}) to $p_{l_1}$ is selected as $p_{l_2}$ (thus, in this case, the set $G$ in (\autoref{eq2}) should be considered the set $G_1 = \left\{p_{l_1}\right\}$, consisting of a single element $p_{l_1}$).
Then, as $p_{l_3}$, the element closest among the remaining elements to the set $G_2 = \left\{p_{l_1}, p_{l_2}\right\}$ consisting of two already selected elements is selected, and so on.
At an arbitrary $k$ - th step of the algorithm, the element $p_{l_k}$ is selected that is closest in terms of magnitude (\autoref{eq2}) among the elements not yet selected to the set $G_{k-1} = \left\{p_{l_1}, p_{l_2},\ldots, p_{l_{k-1}}\right\}$, consisting of the elements selected in the previous steps $k-1$.
If the second property of clusters specified above is true, then the sequence of elements to build should behave as follows.
Elements from the same cluster must go first.
Only after all the elements from one cluster have been exhausted, an element from another cluster appears in the sequence, and then the elements belonging to this second cluster must go on until they are completely exhausted. etc.
Thus, in accordance with the above representation, the sequence of elements built by the algorithm consists of sections, each of which includes all the elements belonging to the same cluster.
Now, to solve the clustering problem, you only need to split the entire sequence of elements into such sections.
For this purpose, a sequence of corresponding numbers is constructed simultaneously with the sequence of elements$K(G_1, p_{l_2}), K(G_2, p_{l_3}),\ldots,K(G_M, p_{l_M})$ - this second sequence is called the "spectrum", hence the name of the algorithm.
In accordance with the properties of the clusters range should behave in the following way.
First, while the first sequence contains elements from the same cluster, the spectrum must contain sufficiently large numbers.
But as soon as an element from a new cluster appears in the first sequence, which is characterized by relatively small coupling values with all the previous elements, a small number should appear in the spectrum.
Then all the numbers in the spectrum should increase again (since the first sequence contains elements from a new aggregate) and be set at a sufficiently high level.
We maintain this level until an element from the new cluster appears.
The element should lead to a drop in the corresponding number in the spectrum, and so on.
In practice, two rules are used for splitting the sequence into sections.

\begin{description}
\item[Rule 1.] 
Choose $L-1$ of such elements $p_{l_k}$ that the numbers $K(G_{k-1}, p_{l_k})$ are the smallest in the spectrum.
These elements, along with $p_{l_1}$, are considered the initial elements for the next sections.

\item[Rule 2.] 
From the spectrum, for each element $p_{l_k}$, the number is calculated

\begin{equation}\label{eq3}
\frac{K(G_{k-2}, p_{l_{k-1}}) - K(G_{k-1}, p_{l_{k}})}{K(G_{k-2}, p_{l_{k-1}})}
\end{equation}

showing the relative change in the spectrum.
Then we select $L-1$ elements $p_{l_k}$, which correspond to the maximum values of the values (\autoref{eq3}), i.e. the largest relative drops in the spectrum.
The selected elements, along with the $p_{l_1}$ element, are considered the starting elements for the next sections.
\end{description}

\section{List of references}
\begin{enumerate}
\item V. Qazvinian, A. Rassoulian, M. Shafiei, J. Adibi. A Large-Scale Study on Persian Weblogs. In \emph{Proceedings of 12th International Joint Conference on Artificial Intelligence, Workshop of TextLink2007 (2007)}.
\item G. Szabo, B.A. Huberman. Predicting the popularity of online content. In \emph{Complexity Digest 2009.02}.
\item X. Guo, D. Vogel, Z. Zhou, Xi Zhang, H. Chen. Chaos Theory as a Model for Interpreting Weblog Traffic. In \emph{Proceedings of the 41st Annual Hawaii International Conference on System Sciences}. 2008.
\item Q. Mei, C.Liu, H.Su, C.Zhai. A Probabilistic Approach to Spatiotemporal Theme Pattern Mining on Weblogs. In \emph{Proceedings of the 15th international conference on World Wide Web}. Edinburgh, UK, 2006.
\item Meishan Hu , Aixin Sun , Ee-Peng Lim. Comments-oriented document summarization: understanding documents with readers' feedback. In \emph{Proceedings of the 31st annual international ACM SIGIR conference on Research and development in information retrieval}. July 20-24, 2008, Singapore, Singapore.
\item F. Duarte, B. Mattos, A.Bestavros, V.Almedia, and J.Almedia. Traffic Characteristic and Communication Patterns in Blogosphere. In \emph{Proceedings of the 1st International Conference on Weblogs and Social Media (ICWSM'06)}. Boulder, Colorado, USA, March 2007.
\item A. Kalterbrunner, V. G\'{o}mez, A.Moghnieh, R.Meza, J.Blat, and V. L\'{o}pez. Homogeneous temporal activity patterns in a large scale online communication space. In \emph{Proceedings of the BIS 2007 Workshop on Social Aspects of the Web (SAW 2007)}. Poznan, Poland, 2007.
\item A. Kalterbrunner, V. G\'{o}mez, V. L\'{o}pez. Description and Prediction on Slashdot Activity. In \emph{Proceedings of the 2007 Latin American Web Conference (LA-WEB 2007)}. Santiago, Chile October 31-November 02 2007.
\item G. Mishne and N. Glance. Leave a reply: An Analysis of Weblog Comments. In \emph{WWW2006, 3rd Annual Workshop on the Weblogging Ecosystem}. Edinburgh, UK, 2006.
\item D.Shen, J.-T. Sun, Q.Yang, Z.Chen. Latent Friend Mining from Blog Data. In \emph{Proceedings of the Sixth International Conference on Data Mining}. Washington, DC, USA, 2006.
\item B. Berendt and R. Navigli. Finding your way through blogspace: using semantics for cross-domain blog analysis. In \emph{AAAI Symposium on Computational Approaches to Analyzing Weblogs}. Stanford, 2006.
\item A. Qamra, B.Tseng, E.Y. Chang. Mining Blog Stories Using Community-Based and Temporal Clustering. In \emph{Proceedings of the 15th ACM international conference on Information and knowledge management}. Arlington, Virginia, USA, 2006.
\item A. Kritikopoulos, M. Sideri, I. Varlamis. BlogRank: Ranking Weblogs Based on Connectivity and Similarity Features. In \emph{Proceedings of the 2nd international workshop on Advanced architectures and algorithms for internet delivery and applications}. Pisa, Italy, 2006.
\item H. Ning, W. Xu. Incremental Spectral Clustering With Application to Monitoring of Evolving Blog Communtites. In \emph{International Conference on Knowledge Discovery and Data Mining}. San Jose, California, USA, 2007.
\item L. Araujo, J. J. Merelo. A Genetic Algorithm for Dynamic Modelling and Prediction of Activity in Document sequences. In \emph{Genetic And Evolutionary Computation Conference}. London, UK, 2007.
\item J. Leskovec, M. McGlohon, C. Faloutsos, N. Glance, and M. Hurst. Cascading Behavior in Large Blog Graphs. Technical Report CMU-ML-06-113, October 2006.
\item E. Keogh. Exact Indexing of Dynamic Time Warping. In \emph{Proceedings of the 28th VLDB Conference}. Hong Kong, China, 2002.
\item Byoung-Kee Yi, H. V. Jagadish, Christos Faloutsos. Efficient Retrieval of Similar Time Sequences Under Time Warping. In \emph{Proceedings of the Fourteenth International Conference on Data Engineering}. Orlando, FL, USA, 1998
\end{enumerate}


\begin{acks}
To Robert, for the bagels and explaining CMYK and color spaces.
\end{acks}

\bibliographystyle{ACM-Reference-Format}
\bibliography{sample-base}

\newpage
\appendix
\section{Overview of clustering methods}
There are several typical clustering methods: K-means, Graphs, and Spectrum (see Appendix 6.3).

\paragraph{The K-means algorithm.}
 The K-means clustering algorithm consists of four steps.
 \begin{enumerate}
 \item Sets the number of clusters $k$ to be formed from objects in the original selection.
 \item Randomly selected $k$ records of the original sample. 
 They serve as the initial cluster centers-centroids.
 \item Filling the resulting$ k $ clusters: for each of the remaining objects, the proximity to the centroid of the corresponding cluster is determined.
 After that, the object is assigned to the cluster whose centroid it is closest to.
 \item For each of the new clusters, the centroid-the object closest to all vectors from this cluster-is calculated anew.
 \end{enumerate}

Steps 3 and 4 are performed until the cluster boundaries and centroid locations no longer change from iteration to iteration, i.e. the same set of records remain in each cluster for each iteration.
The k-means algorithm gained its popularity due to the following properties.
First, it is a moderate computational cost that increases linearly with the number of objects in the original data sample.
Second, the results do not depend on the order of entries in the original sample.
One of the disadvantages of $k$ - means is the strong sensitivity to the choice of initial approximations of centroids.
Random initialization of centers in the first step can lead to poor clustering. 
Clustering may also be inadequate if the number of clusters is initially incorrectly guessed.
It is necessary to perform clustering for different values of $k$, and choose the value at which a sharp improvement in the quality of clustering is achieved for a given functional.

\paragraph{Graph algorithms.}
An extensive class of clustering algorithms is based on the representation of a sample as a graph.
The vertices of the graph correspond to the selection objects. and edges are pairwise distances between objects $\rho_{ij} = \rho(x_i, x_j)$, where $\rho(x_i, x_j)$ is the distance between objects, and $x_i, x_j$ are digital descriptions of objects $i$ and $j$ in the clustered sample.
The advantage of graph clustering algorithms is their visibility, relative ease of realization, and the ability to make various improvements based on simple geometric considerations.
One type of graph algorithms is \emph{algorithm for selecting connected components}.
The parameter $R$ is set and all edges $(i,j)$ for which $\rho_{ij} > R$  are deleted in the graph.
Only the closest pairs of objects remain connected.
The idea of the algorithm is to find a value $R\in [\min \rho_{ij}, \max \rho_{ij}]$ at which the graph falls apart into several connected components.
The connected components found are be clusters.
Note two disadvantages of this algorithm.
Limited applicability.
The algorithm for selecting connected components is most suitable for selecting clusters such as thickenings and ribbons.
The presence of a sparse background or ``narrow jumpers`` between clusters leads to inadequate clustering.
Poor management of the number of clusters.
For many applications, it is more convenient to set the number of clusters or some ``clustering clarity`` threshold rather than the $R$ parameter.
It is rather difficult to control the number of clusters using the $R$ parameter.
You have to perform clustering several times for different $R$.

\section{Figures}
\begin{figure}[ht]
\centering
\includegraphics[height=4.7cm, width=\columnwidth]{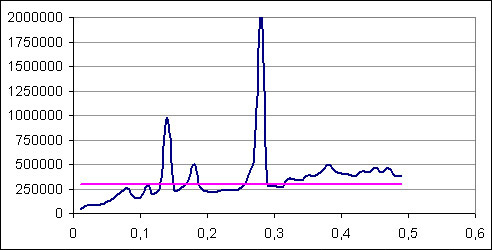}
\caption{\textbf{Time delay per post. The average.}}
\label{post-penalty-mean}
\end{figure}

\begin{figure}[ht]
\centering
\includegraphics[height=4.7cm, width=\columnwidth]{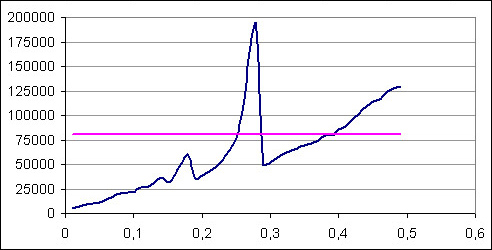}
\caption{\textbf{Time delay per post. Median.}}
\label{post-penalty-median}
\end{figure}

\begin{figure}[ht]
\centering
\includegraphics[height=4.7cm, width=\columnwidth]{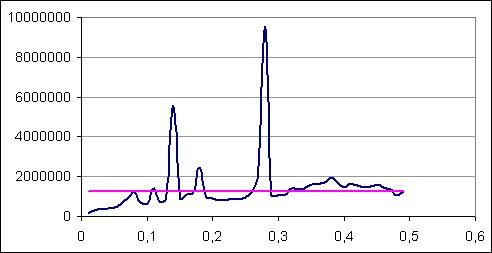}
\caption{\textbf{Time delay per post. Quantile 95\%.}}
\label{post-penalty-q95}
\end{figure}

\begin{figure}[ht]
\centering
\includegraphics[width=\columnwidth]{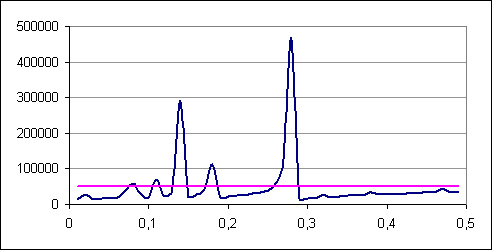}
\caption{\textbf{Time delay for comment. The average.}}
\label{comment-penalty-mean}
\end{figure}

\begin{figure}[ht]
\centering
\includegraphics[width=\columnwidth]{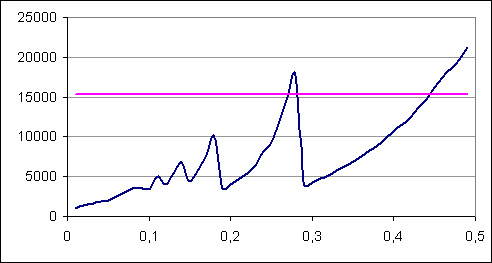}
\caption{\textbf{Time delay for comment. Median.}}
\label{comment-penalty-median}
\end{figure}

\begin{figure}[ht]
\centering
\includegraphics[width=\columnwidth]{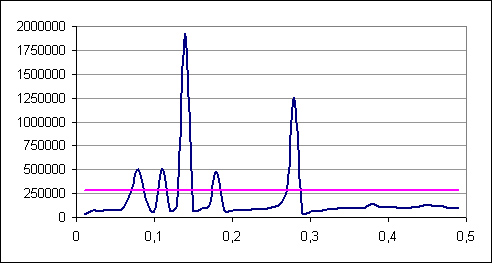}
\caption{\textbf{Time delay for comment. Quantile 95\%.}}
\label{comment-penalty-q95}
\end{figure}

\begin{figure}[ht]
\centering
\includegraphics[width=\columnwidth]{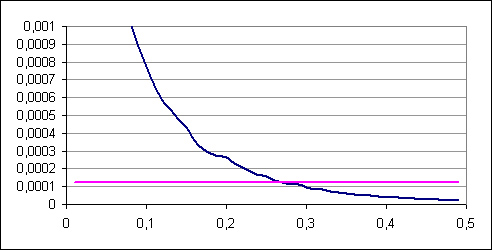}
\caption{\textbf{Relative intensity. The average.}}
\label{relative-intensity-mean}
\end{figure}

\begin{figure}[ht]
\centering
\includegraphics[width=\columnwidth]{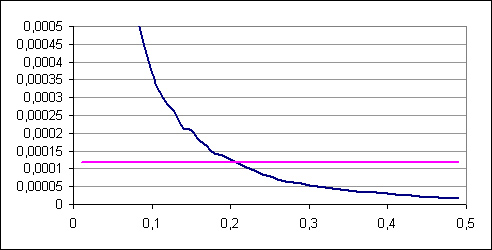}
\caption{\textbf{Relative intensity. Median.}}
\label{relative-intensity-median}
\end{figure}

\begin{figure}[ht]
\centering
\includegraphics[width=\columnwidth]{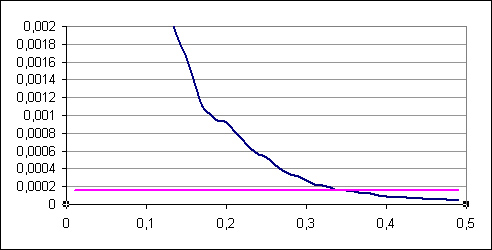}
\caption{\textbf{Relative intensity. Quantile \%.}}
\label{relative-intensity-q95}
\end{figure}

\begin{figure}[ht]
\centering
\includegraphics[width=\columnwidth]{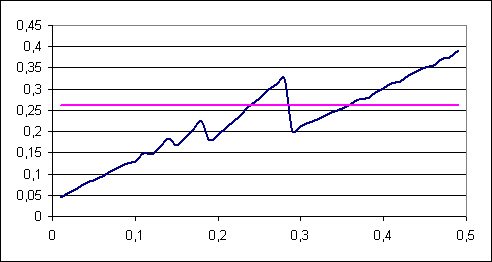}
\caption{\textbf{Probability. The average.}}
\label{probability-mean}
\end{figure}

\begin{figure}[ht]
\centering
\includegraphics[width=\columnwidth]{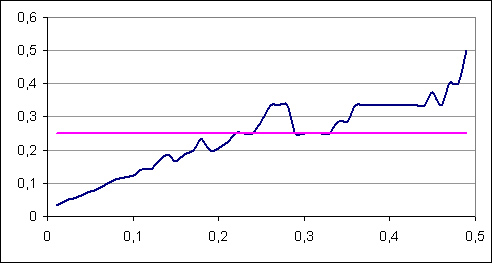}
\caption{\textbf{Probability. Median.}}
\label{probability-median}
\end{figure}

\begin{figure}[ht]
\centering
\includegraphics[width=\columnwidth]{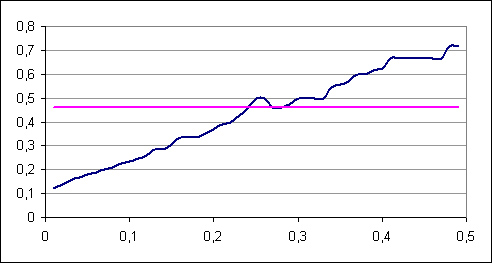}
\caption{\textbf{Probability. Quantile 95\%.}}
\label{probability-q95}
\end{figure}

\begin{figure}[ht]
\centering
\includegraphics[width=\columnwidth]{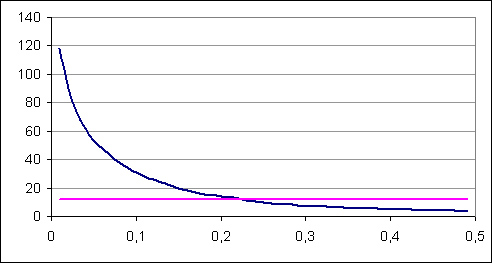}
\caption{\textbf{Absolute intensity. The average.}}
\label{absolute-intensity-mean}
\end{figure}

\begin{figure}[ht]
\centering
\includegraphics[width=\columnwidth]{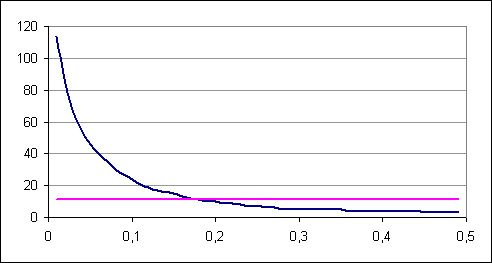}
\caption{\textbf{Absolute intensity. Median.}}
\label{absolute-intensity-median}
\end{figure}

\begin{figure}[ht]
\centering
\includegraphics[width=\columnwidth]{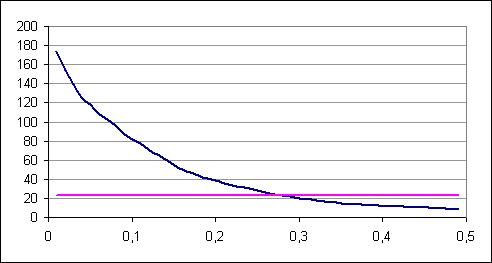}
\caption{\textbf{Absolute intensity. Quantile 95\%.}}
\label{absolute-intensity-q95}
\end{figure}

\section{The Blogosphere}
\subsection{Description of the subject area}
A separate and fast-growing component of all the information available on the Internet is information located in the so-called \emph{blogosphere} - an interconnected network of \emph{blogs}. Just like on regular websites, blogs combine text information with multimedia content, as well as links to other blogs, blog entries, and other sites.
An important feature of most blogs is the ability for readers to leave comments, which themselves become an internal part of the blog and can cause comments from other readers, as well as influence the appearance of new entries in this blog or other blogs.
In this way, the blog network opens up opportunities for multiple relationships.
Even if we consider a separate blog, it is a snapshot of the interactive one-to-many relationship between the blog author and his readers, not just a binary relationship.
A special feature of any blog is how its content changes over time.
Regular web pages are mostly static, but they are subject to arbitrary changes (adding or removing information) that are difficult to track.
The content of the blog changes according to the already established principle-the natural addition of new posts or comments related to the blog, which gives the blog a hierarchical structure and this structure only increases over time.

\paragraph{The size of the blogosphere}
Over the past few years, you can see a significant increase in the size of the blogosphere.
In 2002, Newsweek estimated the number of blogs at half a million, attributing the" explosion " of the blogosphere to the site's emergence Blogger.com (now one of Google's services).
As of November 2006, the blogosphere has reached an incredible size of 60 million blogs, having increased 120 times in 4 years.
All major Internet search engines like Google, Yahoo, and MSN support blog search services as well.
In addition, there are several smaller search engines (Technorati, BlogPulse, Tailrank, and BlogScope) who specialize in searching the blogosphere and exploring the relationships between bloggers.
The table below shows the approximate amount of data used by these search engines.
Among the major search engines, specialized search for blogs have on Google.
The latter also has a specialized search for comments to posts.
\\
\\
\begin{tabular}{|p{0.3\linewidth}|p{0.32\linewidth}|p{0.32\linewidth}|}
\hline
\bf Source & \bf Number of indexed blogs (million) & \bf Number of indexed posts (million)\\
\hline
BlogScope & 33,35 & 646\\
\hline
BlogPulse  & 103,61 & \\
\hline
Technorati   & 133 & \\
\hline
Yandex  & 13,2 & 488\\
\hline
\end{tabular}
\\
\\
\\
In [1], a large-scale study of the Persian blogosphere was conducted, based on which it is possible to estimate the total number of comments in indexed blog entries.
In [1], it is verified that on average each blog post has 3.6 comments.

\paragraph{Comparative growth of the blogosphere}
Below is a comparative growth of the blogosphere by various indexers. \\
\textbf{BlogPulse} (getting started-May, 2004)\\
Dynamics of indexed blogs: \\
10 million in April 2005 \\
12 million in June 2005 \\
100 million January 14, 2009 \\
Over 103 million as of February 25, 2009.\\

\textbf{Technorati} (start of work - 2002)

\begin{figure}[ht]
\centering
\includegraphics[width=\columnwidth]{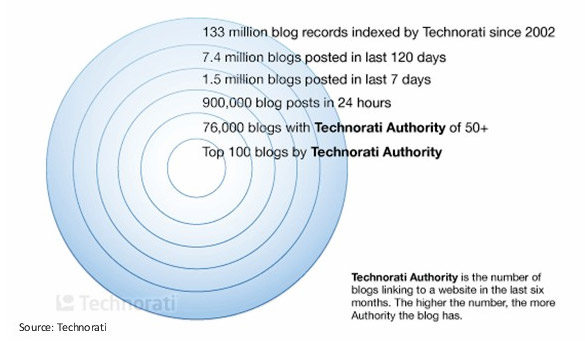}
\caption{Statistics of indexed blogs of the Technorati service}
\label{fig.0}
\end{figure}

\subsection{Literature review}
The appearance of a large number of blogs has provided extensive material for analysis.
Initially, the main work on blogs was aimed at studying the structure of the blogosphere and blogger communities.
For example, [10] discusses the problem of detecting "hidden" friends among bloggers.
Friends are considered bloggers who discuss similar topics in their posts, "hidden" are considered bloggers who did not put links to each other in their posts.
Similar blogs are identified directly both by the content of posts and by the distribution of topics in these posts.
Several methods were considered for determining the proximity of bloggers to each other. The first method considered similarity in the frequency of words used (minus stop words):

\begin{equation*}
    S(i,j) = \frac{\sum_k{n_{ik}n_{jk}}}{\sqrt{\sum_k{n_{ik}^2 \sum_k{n_{jk}^2}}}}
\end{equation*}

where $n_ik$ is the frequency of the k-th word in the I-th blog. 
The second method was based on the analysis of the blogger's post topics:

\begin{equation*}
    D(i,j) = \sum_{t=1}^T{\theta_{it} \log{\frac{\theta_{it}}{\theta_{jt}}} + \theta_{jt} \log{\frac{\theta_{jt}}{\theta_{it}}}}
\end{equation*}

where T is the total number of topics, and $\theta_{it}$ is the probability that the i-th blogger has a topic t.
In [11], methods of semantic analysis of entries for determining blog topics are proposed, as well as grouping (defining communities) of blogs by these topics.
The paper [16] examines the link structure of blogs, citation levels, and the influence of blogs on each other.
As a result, the law of changes in the popularity of old blogs depending on time was established and the main elements of the topological structure of blog communication with each other were proposed.
In [12], a tree of blog connectivity is built based on links in posts to other blogs and communities of bloggers are identified who participate in the discussion of common topics with their posts.
In [13], a ranked structure of the blogosphere is compiled based on the link graph and the modified PageRank algorithm.
The blog's significance level (BlogRank) was calculated using the following formulas:

\begin{equation*}
    B(A) = (1-E) + E[FN(U_1 \to A)B(U_1) + … + FN(U_n \to A)B(U_n)]
\end{equation*}

where $B(A)$ Is The blogrank of the blog $A$, $B(U_i)$ - BlogRank of the blog $U_i$ that links to $A$, $E \in (0,1)$ (usually 0.85), $FN(U_n \to A)$ - the probability that a user who is on the blog $U_n$ goes to the blog $A$.

\begin{equation*}
    FN(U_{z \to j}) = \frac{F_{z \to j}}{F_{z \to x}} 
\end{equation*}

where $F_{z \to x} = L_{z \to x} + w_T*T_{z \to x} +  w_U*U_{z \to x} + w_N$

$L$ - number of links from blog $z$ to blog $j$\\
$T$ - number of common labels (categories) between $z$ and $j$\\
$U$ - the number of users who leave entries in both $z$ and $j$\\
$N$ - number of links from $z$ and $j$ to the shared news URL\\
$w_T$, $w_U$, $w_N$ - weights for using the corresponding parameters ($w_T=2$, $w_U=1$, $w_N=3$)

Work [14] offers a method for clustering blogs by content.
A significant difference from other similar methods is that it allows you to work with a large amount of data in real time.
Moreover, it takes into account the fact that the blog content can change regularly - adding, deleting, and changing blog entries, which is quite critical for the web environment.
In [15], an algorithm is proposed for modeling the time of receipt of new documents from the General information sequence, which can be compared with the exact time of receipt, such as, for example, news, emails, blog entries.
A finite state machine is used to model the data sequence.
In the simplest case, it is assumed that the data source has only two States - with a high generation rate and with a low one.
An exponential probability density function is also used to model the waiting time between incoming documents:

\begin{equation*}
    f(x) = {\alpha}e^{-x} \mbox{ , }\alpha > 0
\end{equation*}

In the $q_0$ state, the source generates documents with a low frequency, which increases the value of the probability density function: $f_0(x) = {\alpha_0}e^{-\alpha_{0}x}$, in the $q_1$ state, the source generates documents with a higher frequency: X
$f_1(x) = {\alpha_1}e^{-\alpha_{1}x}$, $\alpha_1 > \alpha_0$. 
It is also assumed that the probability of changing the state of the automaton p, between two consecutive documents, does not depend on the previous document generation and state changes.
Now you can calculate the probability of a sequence $q = {q_1 \ldots q_n}$ of changing States associated with a sequence $x = {x_1 \ldots x_n}$ of time intervals between $n+1$ documents.
A sequence of States that maximizes the probability of $P(q\mid x)$ minimizes the penalty function $c(q\mid x) = -\ln{P(q\mid x)}$:

\begin{equation*}
    c(q\mid x) = b\ln {\frac{1-p}{p}} +\sum_{t=1}{-\ln{ f_{qt} (x_t) }}
\end{equation*}

A two-state automaton is extended to an infinite number of States, and it is shown that the infinite one can be replaced by an automaton with k-States.

\begin{equation*}
  c(q\mid x) = \sum_{t=0}{\tau(q_t,q_{t+1})} + \sum_{t=1}{-\ln{(\alpha_{qt}e^{-\alpha_{qt}x_t})}}  
\end{equation*}

\begin{equation*}
\tau(q_t,q_{t+1}) = \frac{(q_{t+1} - q_t)}{\ln E} \mbox{ , if } q_{t+1} > q_t \mbox{ , else } \tau(q_t,q_{t+1}) = 0
\end{equation*}

\begin{equation*}
\alpha_q = \frac{n}{T} s^q \mbox{ , } s = e^{\frac{\ln{\alpha_E} - \ln{n} + \ln{T}}{T}}
\end{equation*}

Classical dynamic programming methods are limited by memory and efficiency requirements.
Based on the results of experiments, the proposed algorithm can improve these indicators.
In [2], we propose a method for accurately predicting the long-term popularity of online content by analyzing user traffic at an early stage of content placement.
The analysis was carried out on the example of two services - YouTube and Digg.
Moreover, due to the different types of published materials, to predict popularity during the month, the first one required 10 first days of traffic analysis, and the second only two hours.
Three methods were used to predict popularity:

\subparagraph{1. Linear regression on a logarithmic scale.}
Defining regression parameters on a training sample:

\begin{equation*}
    LSE^* = \sum_c{r_c} = \sum_c{\left({\beta_0(t_i) + \ln{N_c(t_i)} - \ln{N_c(t_r)}}\right)^2}
\end{equation*}

where $N_c(t_r)$  - popularity of the material $c$ at time $t_r$
The prediction of popularity:

\begin{equation*}
   N_s(t_i, t_r) = e^{\ln{N_s(t_i)} + \beta_0(t_i) + \frac{\sigma^2}{2}} \mbox {, where } \sigma = var(r_c) 
\end{equation*}

\subparagraph{2. Minimizing the quadratic error}
As a result, the formula for prediction looks like:

\begin{equation*}
N_s(t_i, t_r) = \alpha(t_i, t_r)N_s(t_i)
\end{equation*}

\begin{equation*}
\alpha(t_i, t_r) = \frac{\sum_c{\frac{N(t_i)}{N(t_r)}}}{\left({\sum_c{\frac{N(t_i)}{N(t_r)}}}\right)^2}
\end{equation*}

The value of $\alpha(t_i, t_r))$ is calculated from the training sample of materials.

\subparagraph{3. Growth model}
Based on the average growth of profiles determined from the training sample:

\begin{equation*}
P(t_0, t_1) = \frac{N_c(t_0)}{N_c(t_1)}
\end{equation*}

\begin{equation*}
N_s(t_r) = \frac{N_s(t_i)}{P(t_i, t_r)}
\end{equation*}

As a result, the logarithmic model showed the smallest absolute error, although the second model should be chosen to minimize the relative error.\\
Work [3] examines the factors that affect user traffic to a blog.
The key points of chaos theory are used to create a model to establish such a dependence of traffic on various factors that cannot be obtained using the usual relationship between factors.
As you can see, many works aim to study and analyze the blogs themselves and their posts.
But in addition to this, there are a large number of comments to the entries.
These comments themselves may be interesting for several reasons. \\
First, despite their informal writing style, comments represent an understanding and "feedback" to the content of a web document. \\
Second, many sites present web documents along with comments to them, comments become an integral part of any blog.\ \
third, the content of comments can better interact with various information extraction services.
For example, many search engines now rank search results by relevance and by date of writing.
Joint analysis of comments for keywords can add additional ranking parameters to the document.
In [5], we study the problem of creating a summary for a web document that also contains comments.
For each comment, three relationships were defined (subject, citation, mention, link) for which comments can be linked.
The importance of each comment is determined: 1) by a General graph made up of three relation graphs; 2) by a tensor method, in which a three-dimensional tensor is formed on the basis of relation graphs.
Then the summary is made up of individual post suggestions and comments, each of which has its own level of importance(significance).
A lot of research in the blogosphere is aimed at studying the blogs themselves, their posts, the author's relationship with each other and authority.
The study of comments began relatively recently.
One of the first and most extensive studies was [9].
It deals mainly with qualitative research - the total volume of comments in blogs, the distribution of the number of comments on blogs (Zipf's law), the quantitative content of comments (the volume of comments, links in them), and the relationship between comments and blog popularity.
The task of detecting user dialogs in comments was also considered. Blogs collected over three weeks with BlogPulse were used as working material.

\begin{figure}[ht]
\centering
\includegraphics[width=\columnwidth]{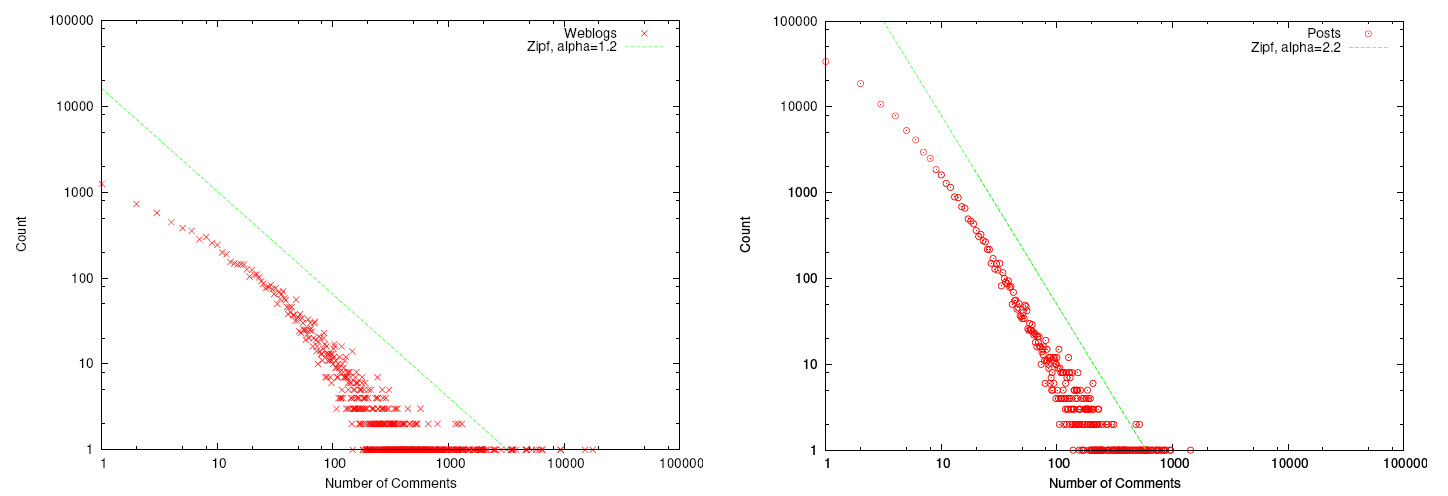}
\caption{\textbf{Power distribution of comments per blog (left) and post (right)}}
\label{fig.1}
\end{figure}

The distribution of comments by day since publication can be seen in [1].
Enter the value  $c_{i,j}$ - the total number of comments left after $j$ days, after posts published on the i-th day.
So $c_{3,10}$ - the total number of comments left ten days later, after the publication of posts on the 3rd day.
Number of comments on day i = $\sum_{1\le j\le i}{c_{i,i-j}}$ for figure.3. a graph that for each day d shows the total number of comments left d days after the corresponding posts, divided by the total number of comments left on the day the post was published.

\begin{figure}[ht]
\centering
\includegraphics[width=\columnwidth]{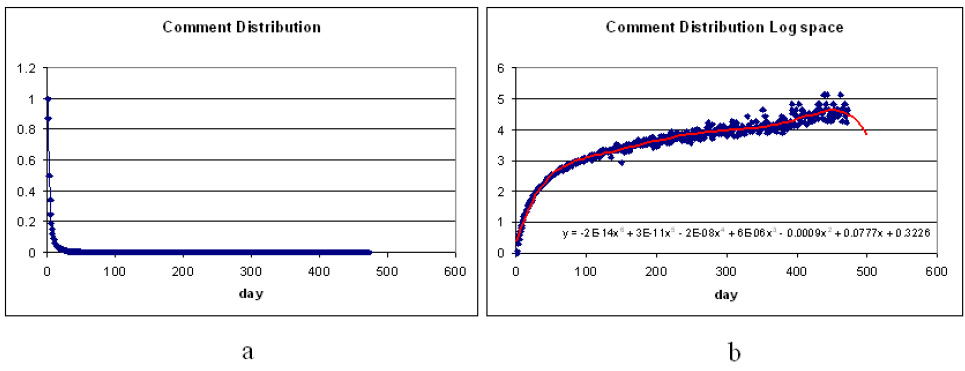}
\caption{\textbf{The distribution of the comments.(a) shows the $\frac{\sum_k{c_{k,day}}}{\sum_k{c_{k,0}}}$ for each day. (b) shows (-log) (a}}
\label{fig.2}
\end{figure}

Data for 400 days was interpolated by a power function:

\begin{align*}
    F(x) &= (-2e-14)x^6 + (3e-11)x^5 - (2e-08)x^4 \\ 
    &+(6e-06)x^3 - 0.0009x^2 + 0.0777x + 0.3226
\end{align*}

\begin{figure}[ht]
\centering
\includegraphics[width=\columnwidth]{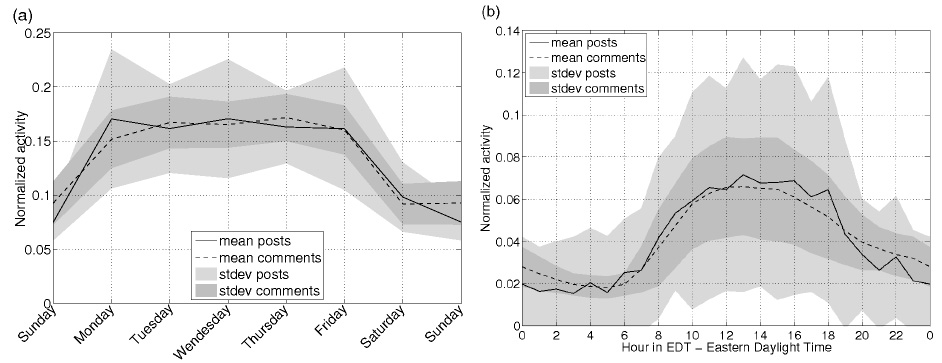}
\caption{\textbf{(a) Weekly and (b) daily activity cycles}}
\label{fig.3}
\end{figure}

Articles [7, 8] consider user activity on a news site slashdot.org
\textbf{Post-Comment-Interval} (PCI) - the time difference between the comment and the posts that the comment belongs to.\\
\ textbf{Inter-Comment-Interval} (ICI) - the difference between two consecutive comments from the same user (regardless of which post was commented out)\\
As can be seen in Figure 4. high and constant activity is observed during the working week.
While daily activity reaches its maximum around one o'clock in the afternoon and minimum around 3-4 o'clock in the morning.
Although the Slashdot site is open to the world, the activity profile corresponds to the American daily routine.

\begin{figure}[ht]
\centering
\includegraphics[width=\columnwidth]{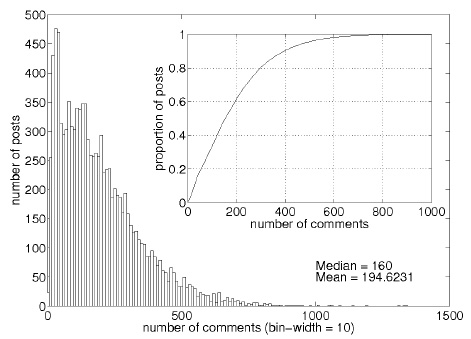}
\caption{\textbf{Histogram of the number of comments per post (the subgraph shows the corresponding distribution function)}}
\label{fig.4}
\end{figure}

The histogram in Fig.5. gives an idea of the number of comments a post can receive.
It can be seen that this distribution, as in [1], also obeys Zipf's law.
Note that half of the posts have more than 160 comments, and some exceed 1000 comments.
From Fig.6. it can be seen that the distribution of comments per user is also subject to Zipf's law.

\begin{figure}[ht]
\centering
\includegraphics[width=\columnwidth]{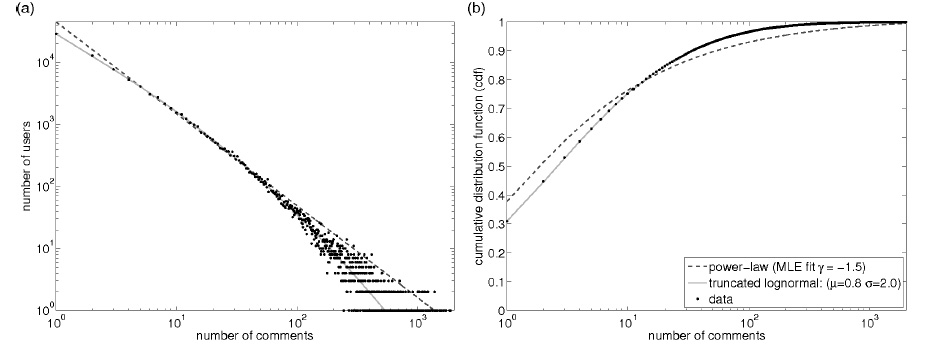}
\caption{\textbf{(a) Distribution of the number of comments per user (b) the corresponding distribution function.)}}
\label{fig.5}
\end{figure}

To predict user activity, you need to create a distribution function for PCI.
The simplest distribution used is log - normal (LN):

\begin{equation*}
f_{LN}(t;\mu,\sigma) = \frac{1}{{t\sigma\sqrt{2\pi}}}~e^{ \frac{{\left(-\ln{t} - \mu\right)}^2} {2\sigma^2} }
\end{equation*}

A double log-normal distribution (DLN) is also used, which is a superposition of two independent log-normal distributions.

\begin{align*}
    f_{DLN}(t;\theta) = k f_{LN}(t;\mu_1,\sigma_1) + (1-k)f_{LN}(t;\mu_2,\sigma_2)
\end{align*}

Here, $\theta = (\mu_1, \sigma_1, k, \mu_2, \sigma_2)$

\paragraph{}
Double lognormal distribution allows you to take into account the daily cycles of user activity.
Each post was found to trigger two waves of comment activity.
The first one starts from the moment the post is published, and the second one starts after the next peak in the total daily user activity.

\begin{figure}[ht]
\centering
\includegraphics[width=\columnwidth]{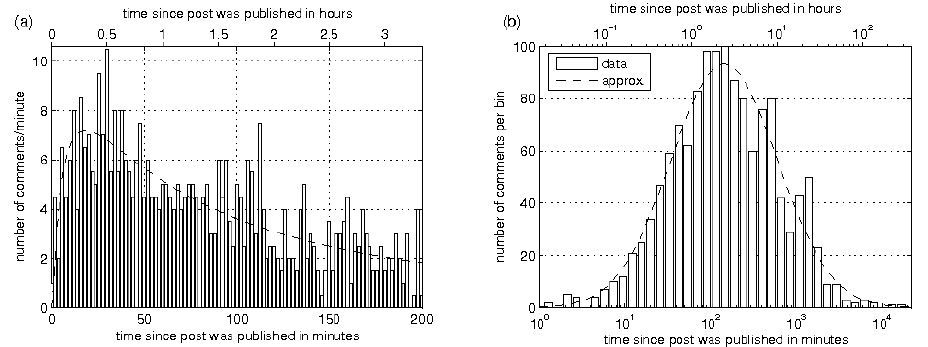}
\caption{\textbf{LN is an approximation of the PCI distribution for a post that received 1,341 comments. (a) comments per minute for the first 200 minutes after the post is published (b) the same on a logarithmic scale.}}
\label{fig.6}
\end{figure}

\paragraph{}
\begin{figure}[ht]
\centering
\includegraphics[width=\columnwidth]{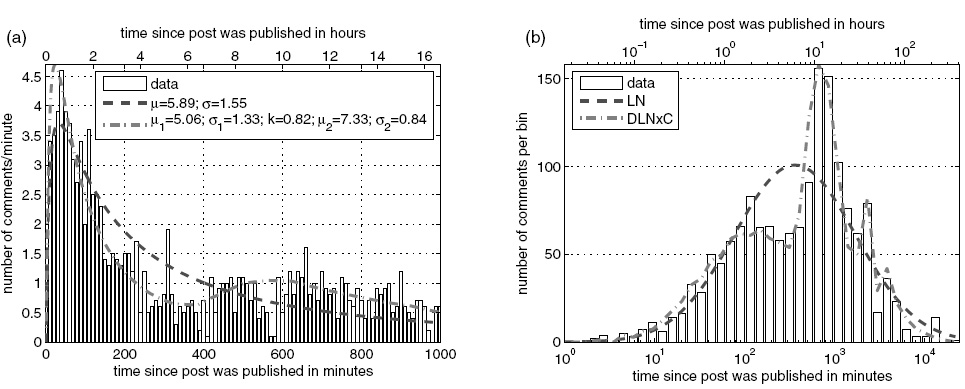}
\caption{\textbf{Comparison of LN(light hatching) and DLN (dark hatching) distributions for a post published at night. Description of graphs as in Fig. 6}}
\label{fig.7}
\end{figure}

Figure 7 shows an example of a graph where the post was published at night, i.e. the two activity waves do not coincide, which can be seen as an approximation using LN and DLN.
Later in this paper, we proposed a method for predicting the activity of fasting.
To do this, we first created prototypes of activity based on the training sample.
A total of 24 prototypes were created, depending on the daily time when the post was published (one for each hour).
Then, taking into account the predefined prototypes, and taking into account the activity in the first few minutes after the post was published, the function of distributing comments for each post was described.

\begin{figure}[ht]
\centering
\includegraphics[width=\columnwidth]{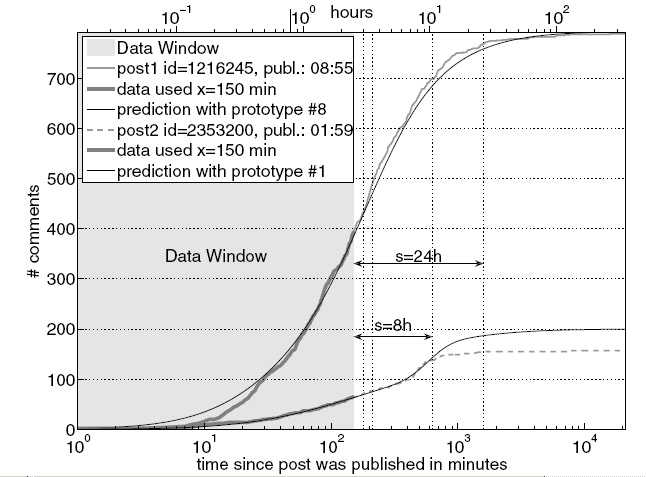}
\caption{\textbf{Two examples of predicting a distribution with predefined activity prototypes.}}
\label{fig: homogeneous6}
\end{figure}

As a result, we have an effective model for predicting the distribution function of comments, but this function is built for each individual post and you need a sufficient number of comments to build the distribution.
Slashdot.org it is a popular news resource with its own community of users, so almost every post has several dozen comments.

\subsection{Blogs. Basic definition}
\textbf{A blog } (English \ emph{blog}, from " \ textbf{web log}", "online journal or event diary") is a website whose main content is regularly added entries, images, or multimedia.
Blogs are characterized by small entries of temporal significance, sorted in reverse chronological order (the last entry is displayed first in the list).
We can call entries posts

\textbf {Bloggers} (bloggers) are people who run blogs.
According to the author's composition, blogs can be personal, group (corporate, club \ldots ) or public (open).
By content - thematic or General.
\textbf {Blogosphere} (from the English \emph{blogosphere}) - a term that refers to the totality of all blogs, as a community or social networks.
There are tens of millions of blogs in the world that are usually closely linked, bloggers read and comment on each other, link to each other, and thus create their own subculture.
The concept of the blogosphere emphasizes one of the main differences between blogs and ordinary web pages and Internet forums: related blogs can form a dynamic global information envelope.
\textbf{Tags} (tags, labels, categories) - keywords that describe the object in question, or relate it to a category.
These are a kind of placemarks that are assigned to an object to determine its place among other objects.
Tags can be assigned to either a blog or a separate post in the blog.
\textbf{Multimedia} (lat. \emph{Multum} + \emph{Medium}) - simultaneous use of various forms of information representation and processing in a single container object.
For example, a single container object (\emph{container}) can contain text, audio, image, and video information, as well as possibly a way to interact with it.

\paragraph{}
\textbf{The Zipf's Law (Zipf)}- an empirical pattern of distribution of the frequency of natural language words: if all the words of the language (or just a long enough text) are ordered in descending order of their frequency of use, the frequency of the nth word in such a list is approximately inversely proportional to its ordinal number n (the so-called rank of this word, see the order scale). For example, the second most used word is about twice as common as the first, the third is three times less common than the first, and so on.

\paragraph{}
\textbf{PageRank} - an algorithm for calculating page authority used by the Google search engine.
PageRank is a numeric value that describes the" importance " of a page in Google.
The more links to a page, the more important it becomes.
In addition, the "weight" of page A is determined by the weight of the link passed by page B.
thus, PageRank is a method for calculating the weight of a page by counting the importance of links to it.
PageRank is one of the auxiliary factors when ranking sites in search results.
PageRank is not the only way to determine a site's position in Google search results, but it is very important.

\section{Dataset}
\begin{figure}[ht]
\centering
\includegraphics[width=\columnwidth]{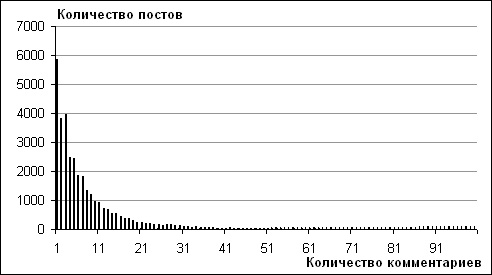}
\caption{\textbf{Distribution of the number of comments by posts in the training sample}}
\label{histogramm}
\end{figure}

\begin{figure}[ht]
\centering
\includegraphics[width=\columnwidth]{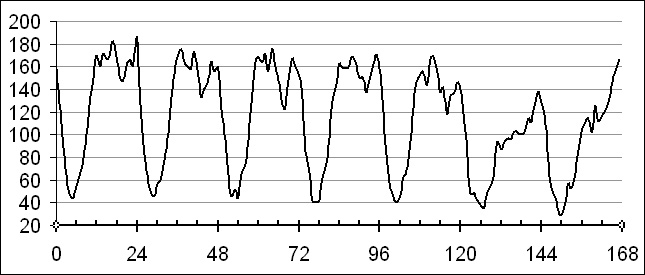}
\caption{\textbf{Daily post publication cycles}}
\label{cyrcadian}
\end{figure}

Figure~\autoref{histogramm} shows the distribution of posts by the number of comments in them.
Figure~\autoref{cyrcadian} shows the daily activity cycles of publishing posts in the training program

\subsubsection{Data with COHORT1}

\begin{figure}[ht]
\centering
\includegraphics[width=\columnwidth]{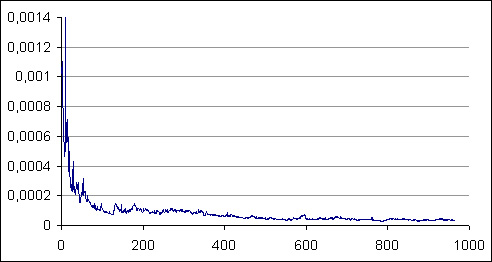}
\caption{\textbf{General view of the spectrum for the first 1000 posts}}
\label{spectre1}
\end{figure}

\begin{figure}[ht]
\centering
\includegraphics[width=\columnwidth]{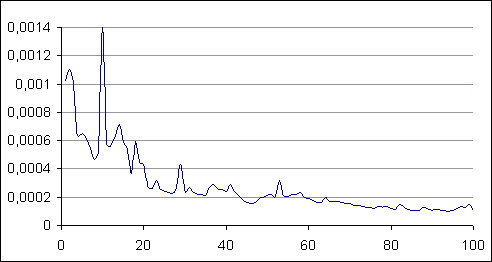}
\caption{\textbf{Spectrum view for the first 100 posts}}
\label{spectre2}
\end{figure}

\begin{figure}[ht]
\centering
\includegraphics[width=\columnwidth]{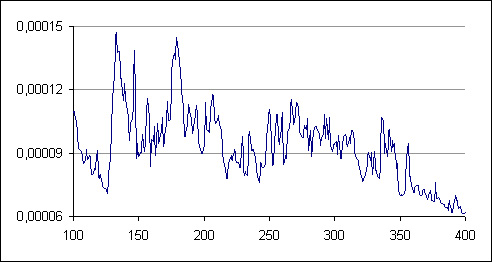}
\caption{\textbf{Spectrum type for posts from 100 to 400}}
\label{spectre3}
\end{figure}

\begin{figure}[ht]
\centering
\includegraphics[width=\columnwidth]{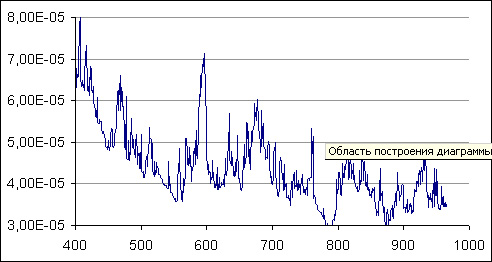}
\caption{\textbf{Spectrum type for posts from 400 to 1000}}
\label{spectre4}
\end{figure}

\begin{figure}[ht]
\centering
\includegraphics[width=\columnwidth]{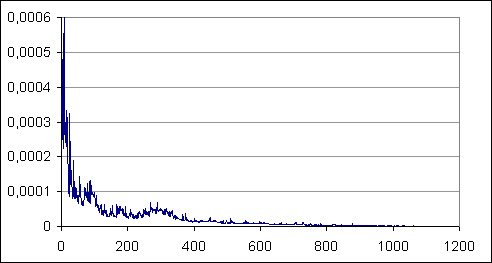}
\caption{\textbf{General view of the spectrum for the first 1200 posts}}
\label{spectre5}
\end{figure}

\begin{figure}[ht]
\centering
\includegraphics[width=\columnwidth]{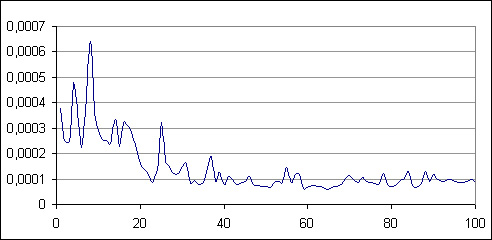}
\caption{\textbf{Spectrum view for the first 100 posts}}
\label{spectre6}
\end{figure}

\begin{figure}[ht]
\centering
\includegraphics[width=\columnwidth]{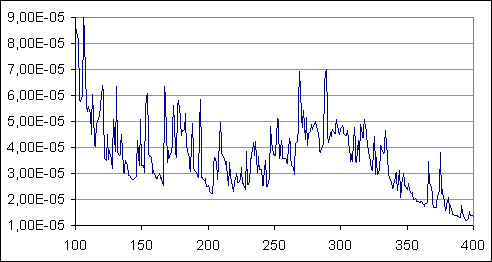}
\caption{\textbf{Spectrum type for posts from 100 to 400}}
\label{spectre7}
\end{figure}

\begin{figure}[ht]
\centering
\includegraphics[width=\columnwidth]{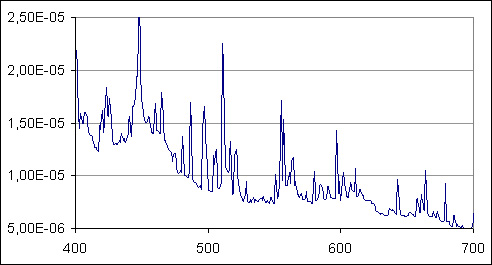}
\caption{\textbf{Spectrum type for posts from 400 to 700}}
\label{spectre8}
\end{figure}

\autoref{spectre1}  shows an example of the spectrum for a set of posts with more than 20 comments, and~\autoref{spectre5} shows the spectrum of posts with more than 40 comments.~\autoref{spectre2},~\autoref{spectre3},~\autoref{spectre4}, and~\autoref{spectre6},~\autoref{spectre7},~\autoref{spectre8} show sections of the spectra at an enlarged scale.

As can be seen from the obtained drawings of the `Spectrum", there are borders between clusters, but they are not clear. Therefore, a proprietary algorithm is used for further clustering.

\subsubsection{DATA with COHORT2}
The values of the algorithm parameters described above, to the following:\\
$T_h$ - threshold value of time $T_h$ = 15 days = 15*86400 seconds,\\
$T_b$ - the base value for mesh, $T_b$ = 1 day = 86400 seconds,\\
$\gamma$ parameter of the grid, $\gamma = 1.04$.

Thus, the time grid with nodes $T_j = \gamma^jT_b$ has the form:

\begin{equation*}
    T^* = (33, 35, \ldots,..., 83077, 86400, 89856,\ldots, 1243887)
\end{equation*}

A total of $\approx$260 nodes were received.
The entire training sample was clustered for each node.
The minimum cluster size is $N = 100$.
The result was about 10,000 clusters.
For each cluster, the optimal time for the comment to appear was calculated, as described above.
As a result, the optimal time for comment appearance was calculated for all clusters and for each grid node.
After calculating all the estimates for the optimal time for the comment to appear, you need to evaluate the quality of the forecast.
To do this, select another 10,000 posts (test sample) with all the known comments and conduct a series of experiments.

\end{document}